\documentclass[runningheads]{llncs}
\usepackage{graphicx}
\usepackage{comment}
\usepackage{color}
\usepackage{times}
\usepackage{epsfig}
\usepackage{bbm}
\usepackage{adjustbox}
\usepackage{tabu}
\usepackage{booktabs}
\usepackage{caption} 
\usepackage{amsmath,amssymb} 
\newif\ifsubmit

  \submittrue

\ifsubmit
\newcommand{\zhe}[1]{}
\newcommand{\ruoyu}[1]{}
\newcommand{\sheng}[1]{}
\else
\newcommand{\zhe}[1]{{\textcolor{blue}{[Zhe: #1]}}}
\newcommand{\ruoyu}[1]{{\textcolor{magenta}{[Ruoyu: #1]}}}
\newcommand{\sheng}[1]{{\textcolor{green}{[Sheng: #1]}}}
\fi

\newcommand{\ie}{\textit{i.e.}}
\newcommand{\eg}{\textit{e.g.}}
\newcommand{\etal}{\textit{et al.}}

\newcommand{\secref}[1]{Section~\ref{#1}}
\newcommand{\figref}[1]{Figure~\ref{#1}}
\newcommand{\tabref}[1]{Table~\ref{#1}}

\usepackage[pagebackref=true,breaklinks=true,letterpaper=true,colorlinks,bookmarks=false]{hyperref}

\begin{document}

\mainmatter

\title{A New Super-Resolution Measurement of Perceptual Quality and Fidelity} 

\author{Sheng Cheng 
\\
{\tt\small chengsheng0210@gmail.com}
\institute{}
\thanks{The work is mentored by Ruoyu Sun and Zhe Hu.}
}

\maketitle

\begin{abstract}
  Super-resolution results are usually measured by full-reference image quality metrics or human rating scores.
  However, these evaluation methods are general image quality measurement, and do not account for the nature of the super-resolution problem.
  In this work, we analyze the evaluation problem based on the one-to-many mapping nature of super-resolution, and propose a novel distribution-based metric for super-resolution.
  Starting from the distribution distance, we derive the proposed metric to make it accessible and easy to compute. 
  %
  Through a human subject study on super-resolution, we show that the proposed metric is highly correlated with the human perceptual quality, and better than most existing metrics. 
  Moreover, the proposed metric has a higher correlation with the fidelity measure compared to the perception based metrics.
  To understand the properties of the proposed metric, we conduct extensive evaluation in terms of its design choices, and show that the metric is robust to its design choices. 
  Finally, we show that the metric can be used to train super-resolution networks for better perceptual quality.
\end{abstract}

\section{Introduction}
%
%
Single image Super-Resolution (SR) is the process of recovering a High-Resolution (HR) image from an observed Low-Resolution (LR) input~\cite{freeman2000learning}.
It is a fundamental problem in computer vision and has a wide range of applications, such as smartphone imaging, medical imaging~\cite{greenspan2009super} and surveillance.
A major challenge is that a low-resolution image may correspond to multiple high-resolution images, thus the problem can be viewed as a one-to-many mapping problem which is an ill-posed problem.
%
As a result, the evaluation of SR methods is quite challenging since we need to consider both the similarity to ground-truth HR images and the perceptual quality. 

The current evaluation metrics can be categorized into two classes:
the first class uses fidelity-based metrics that focuses on the similarity to the reference HR images, and the second class evaluates the perceptual quality .
%
Fidelity-based image quality metrics have been long and widely used for SR tasks, as well as other image and video processing problems.
Commonly used metrics include the Mean Squared Error (MSE), Peak Signal-to-Noise (PSNR) and the Structural SIMilarity (SSIM)~\cite{wang2004image}. 
%
%
These metrics evaluate the similarity between the reconstructed image and the given reference image under a signal fidelity criteria.
However, these metrics typically disagree with the perceptual quality of the Human Visual System (HVS)~\cite{blau20182018}.
Therefore, researchers commonly conduct human subject studies
to further evaluate the reconstructed images, \ie,
collecting human judgements on the perceptual quality of the reconstructed images.
But the human subject study is inconvenient and does not provide instant feedback.

There has been great effort for designing computable metrics that are consistent with the perceptual quality, such as BRISQUE~\cite{mittal2012no} and NIQE~\cite{mittal2012making}.
They have higher correlation with human visual systems than most fidelity based metrics. 
However, these perceptual quality metrics are often biased towards visually pleasing contents and ignore the fidelity, \ie, whether the estimated HR image corresponds to the LR input. 
%
Thus, fake HR images which look authentic but have weak connection
to the LR images can still obtain high scores under these metrics.
In an extreme case, a high score will be given to a HR image of a tree, even if the corresponding LR image is a cat. 
%
%
Therefore, an ideal metric for SR evaluation should meet requirements of both the perceptual quality and the fidelity.

Moreover, an underlying issue for most existing metrics is that they do not consider the one-to-many mapping nature of SR problems.
They are general image quality metrics, and are often instance-based metrics. 
%
This would cause undesirable issues when applying these metrics to super-resolution tasks.
%
Specifically, for an LR input, the instance-based full-reference metrics are biased towards the given reference HR images over other possible
ground-truth HR images.
%
For instance, suppose a LR image $L$ can be generated by downsampling\ each of the two different HR images $H_1$ and $H_2$, \ie, both $(L, H_1)$ and $(L, H_2)$ are possible pairs of LR-HR images.
If $H_1$ is designated as the reference in the evaluation phase, the mapping pair $(L,H_1)$ would lead to perfect reconstruction accuracy, while the mapping pair $(L,H_2)$ would not.
An ideal metric for SR should treat $ (L, H_1) $ and $(L, H_2)$ equally. 

In this work, we analyze the SR evaluation problem based on the nature of SR and propose a full-reference and distribution-based metric.
We show that the proposed metric is a good trade-off between the perceptual quality and the fidelity.
%
We answer a few fundamental questions about the metric:
1) How well does the proposed metric correspond to human visual perception?
2) How robust is the proposed metric?
3) Can the proposed metric be used for improving a SR method?
We first conduct a human subject study on ten SR methods and show that the proposed metric has a high correlation with the human perceptual quality and is higher than most existing metrics.
Second, we demonstrate that the proposed metric is robust with respect to the parameters of its configurations.
Third, we apply the proposed metric as the loss for training SR networks, and show that it would lead to visually pleasing results.
The contributions of the paper are summarized as follows:
\begin{itemize}
    \item We define a distribution-based metric specifically for super-resolution.
    \item We conduct a human subject study to show that the metric 
    correlates well with the perceptual quality.
    \item We study the properties of the proposed metric and show that it is robust.
    \item We show that the proposed metric can be used for training SR networks.
\end{itemize}


\section{Related Work}
In the field of SR, existing methods \cite{dong2016image,kim2016accurate,lai2018fast,haris2018deep,lim2017enhanced,Zhang_2018_ECCV,vasu2018analyzing,Wang_2018_ECCV_Workshops,wang2018fully} pose the problem of evaluating SR performance as assessing the image quality of the super-resolved images.
And most of them use general image quality metrics to evaluate SR performance. 
The commonly used metrics include MSE, PSNR, SSIM~\cite{wang2004image}, IFC~\cite{sheikh2005information}, LPIPS~\cite{zhang2018unreasonable}, BRISQUE~\cite{mittal2012no}, NIQE~\cite{mittal2012making}, Ma~\cite{ma2017learning}, PI~\cite{blau20182018}. 
These metrics can be categorized into two groups: distortion measures and perceptual quality (sometimes referred to as full-reference metric and no-reference metric). 

{\flushleft \bf Distortion measures} (full-reference metrics).
The distortion measures evaluate the generated HR images when the ground-truth is provided, and they are often referred to as full-reference metrics.
The mostly used distortion measure for super-resolution tasks is  MSE, a pixel-wise $\ell_2$ measurement, and the related metric PSNR.
To better explore structural information, Wang \etal \cite{wang2004image} propose the SSIM index that evaluates structural similarities between images.
In \cite{sheikh2005information}, Sheikh \etal propose the IFC metric based on the natural image statistics. 
Although the IFC metric achieves higher correlation with the HVS than MSE and SSIM based on a human subject study on the SR benchmark~\cite{ma2017learning}, the correlation is still not very high. 
Recently, Zhang \etal~\cite{zhang2018unreasonable} propose LPIPS that calculates image similarities on the features extracted from a deep neural network. 
The LPIPS achieves surprisingly good results and is shown to correlate well with the HVS for SR tasks.
However, like other metrics mentioned above, it is designed for general image quality assessment, and does not consider the one-to-many mapping nature of the SR problem. 
%

{\flushleft \bf Perceptual quality.}
The perceptual quality metrics evaluate how likely an image is a natural image with respect to the HVS. 
%
%
The straightforward approaches to evaluate the perceptual quality are to calculate the mean opinion scores of human subject ratings on images.
The recent SR methods adopt such approaches to provide the perceptual quality comparison with other methods, besides the comparison of distortion measures. 
%
The human subject study, although an effective way to reflect the features of the HVS, is expensive to conduct and cannot provide instant feedback. 

To provide convenient and automatic evaluation, great efforts have been made for designing general image quality metrics (\eg, BRISQUE~\cite{mittal2012no}, NIQE~\cite{mittal2012making}). 
These no-reference metrics are usually based on the natural image statistics in spatial, wavelet and DCT domains.
These metrics, being the general image quality assessment, do not specifically design for the SR methods.
Recently, Ma \etal~\cite{ma2017learning} propose a no-reference metric (refered to as Ma) specifically for SR problems, and it is learned from the human rating scores on the SR images from different methods.
The metric Ma has been shown to match well with the perceptual quality of the HVS, and it is used for forming the perception index (PI), which is the default measure for the PIRM challenge on SR~\cite{blau20182018}.
Although the no-reference measures are convenient to use, they do not reflect the signal fidelity for SR tasks (whether an estimated HR image matches the input LR image under a certain downsampling process).

The proposed SR metric in this work is derived from the evaluation problem of the SR problem, thus naturally maintaining the signal fidelity. 
And we show that it also highly correlates with the perceptual quality of the HVS.

{\flushleft \bf Distribution-based metric.}
Few works tackles the SR evaluation with distribution based metrics~\cite{sonderby2016amortised,blau2018perception}.
The recent work~\cite{blau2018perception} points out that the SR metric should be distribution-based instead of instance-based.
However, they only evaluate on the HR space, and miss the point that the SR problem is a mapping problem, where the joint space of input and output should be considered. 
In this work, we propose a metric to properly handle this.
%

\section{Proposed Method}
\label{sec:method}
Given a set of ground-truth LR-HR image pairs, and a set of HR images generated by an SR algorithm, the evaluation task is to design a metric to measure the quality of the generated HR images by comparing with the ground-truth. 

In the following, we put the discussion of HR and LR in patch domain, without loss of generality, instead of the whole images. %
Suppose the random variable $Y$ represents the HR patch, with 
a probability density $p_Y$. 
The random variable $X$ represents the LR patch, with a probability density $p_X$. 
We denote the downsampling process from HR patches to LR patches (the inverse mapping of SR) as $g: Y \rightarrow X$. 
SR is a one-to-many mapping problem: for each $X = x$, there is a set of HR patches $\mathbb{Y}_x = \{y_i | g(y_i) = x\}$ corresponding to $x$.
And we can represent the distribution of HR patches corresponding to a given LR patch $x$ using a conditional distribution $p_{Y|X=x}$. 


\begin{figure*}[ht]
  \centering
  \Large
  \begin{adjustbox}{width=\linewidth}
    \begin{tabular}{ccccc}
    \hspace{-3mm}
    \includegraphics[width=0.433\linewidth]{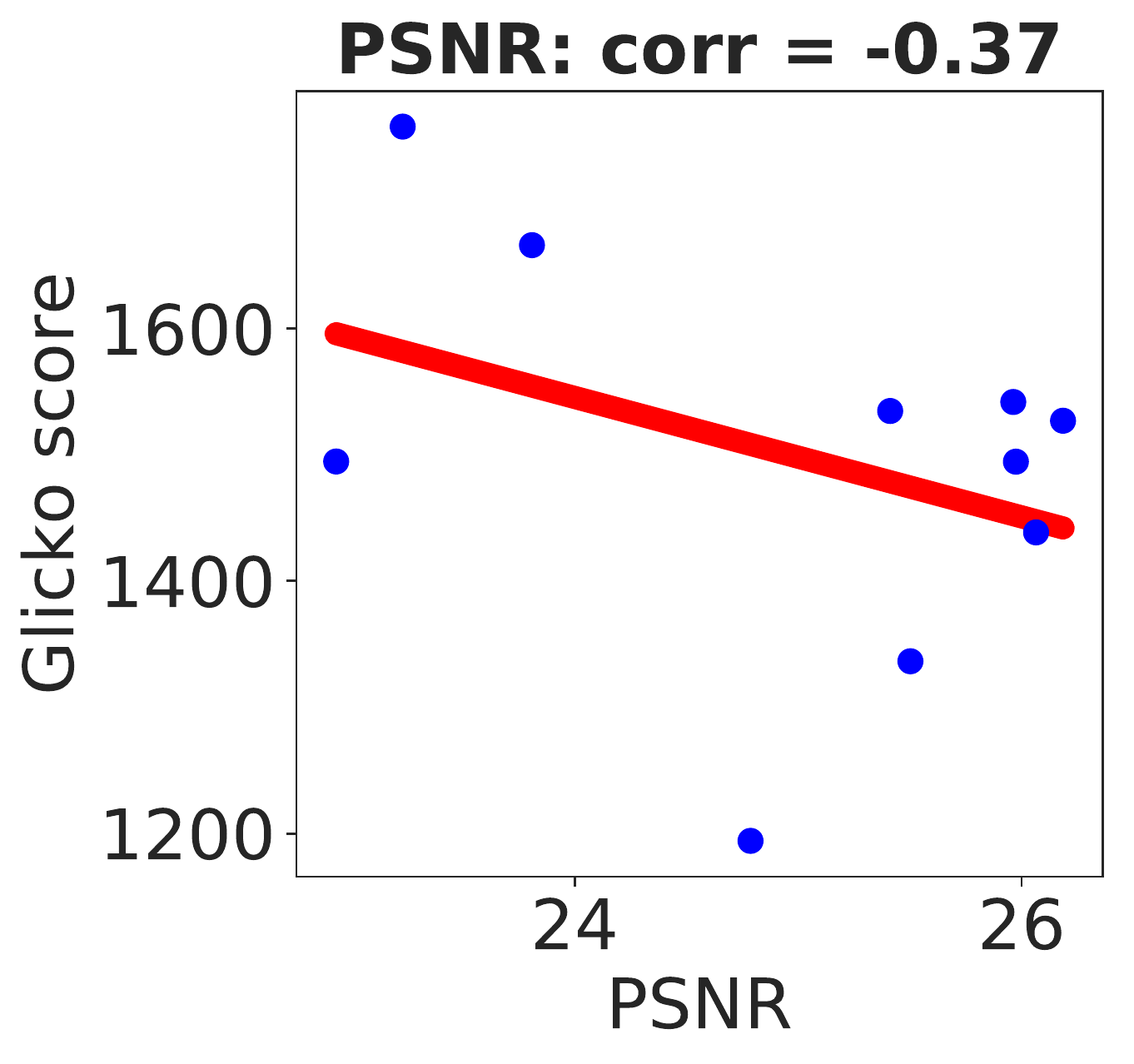} & 
    \includegraphics[width=0.4131\linewidth]{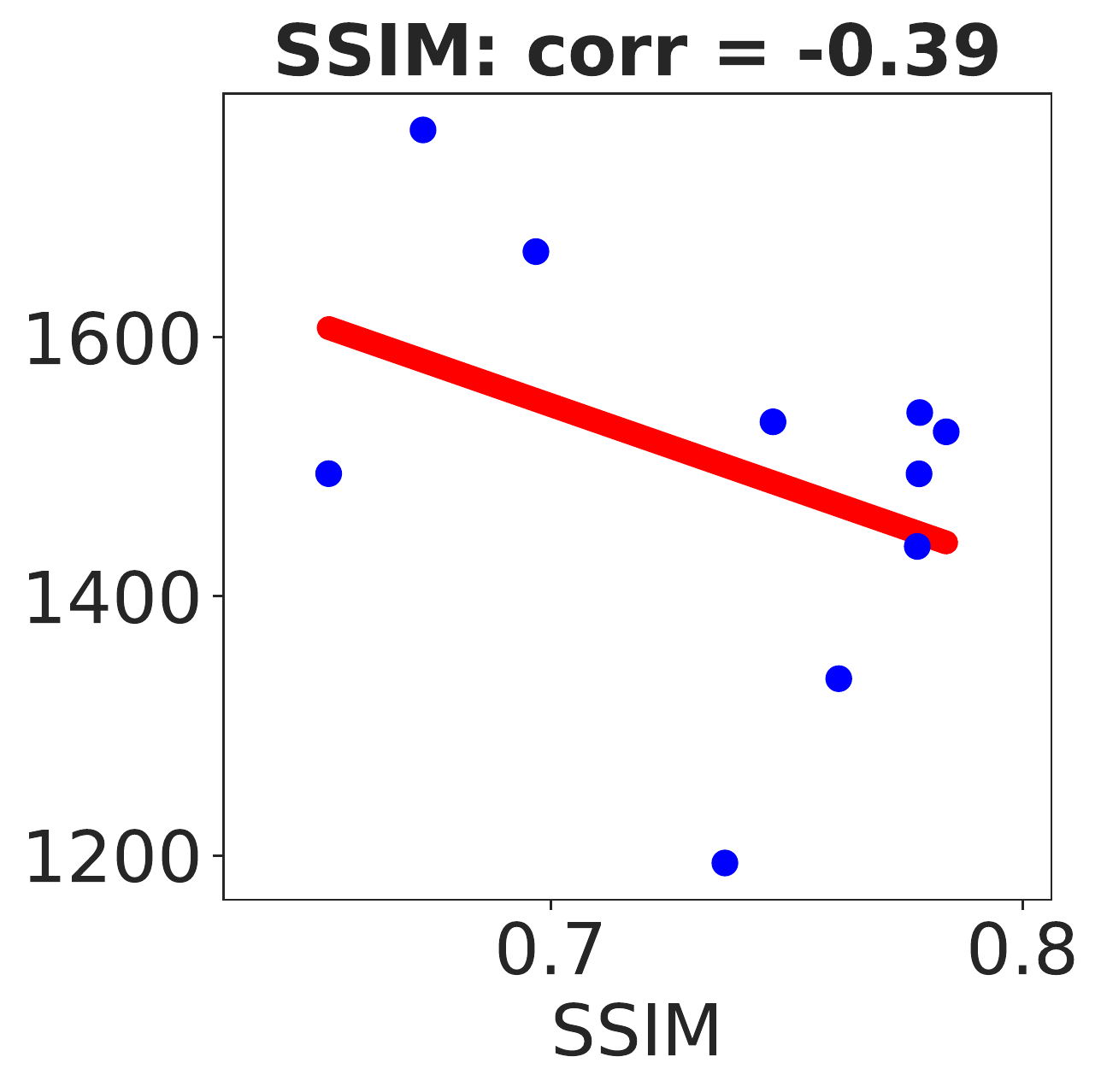} & 
    \includegraphics[width=0.42\linewidth]{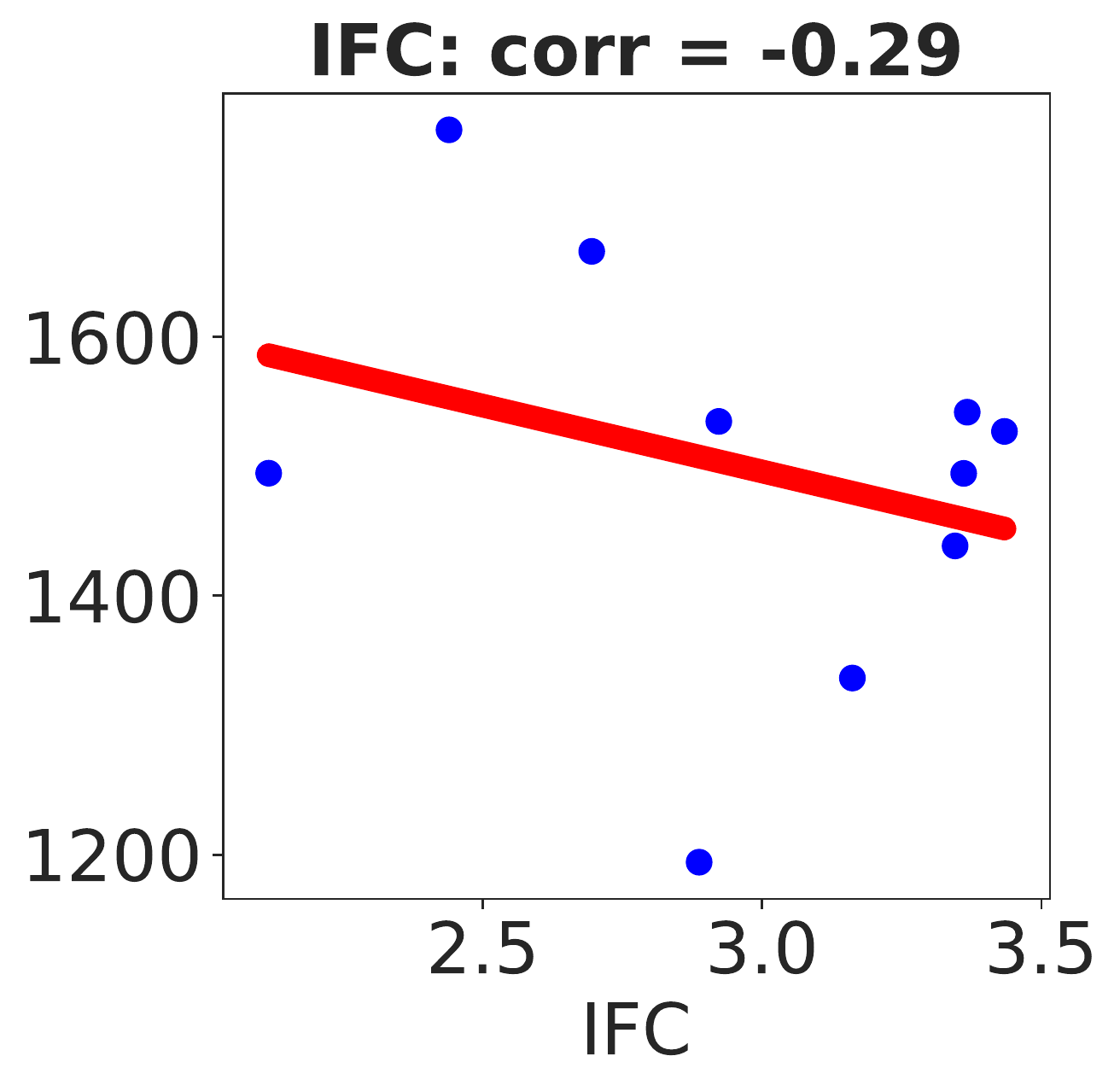} &
      \includegraphics[width=0.4\linewidth]{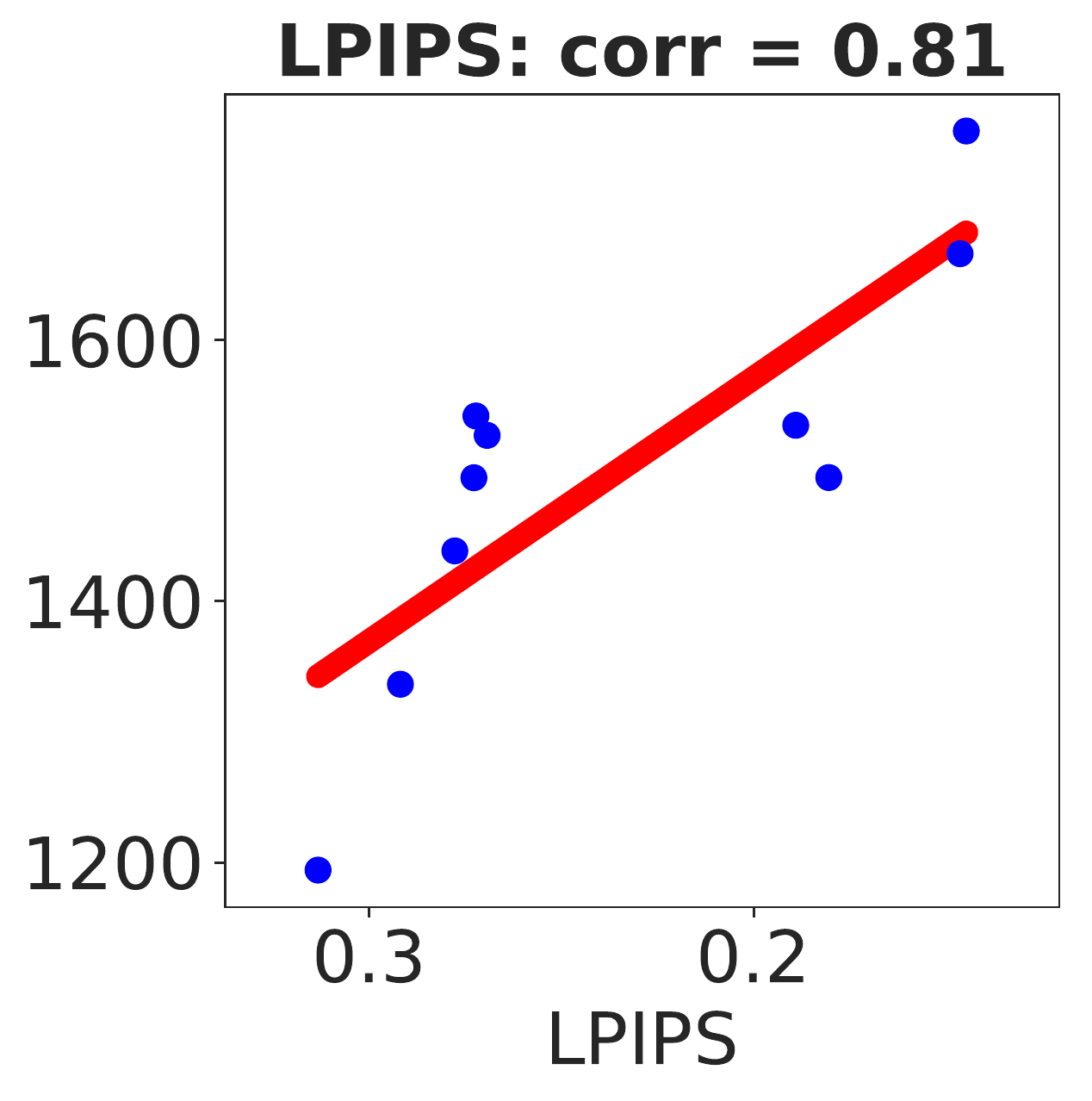} &
      \includegraphics[width=0.4\linewidth]{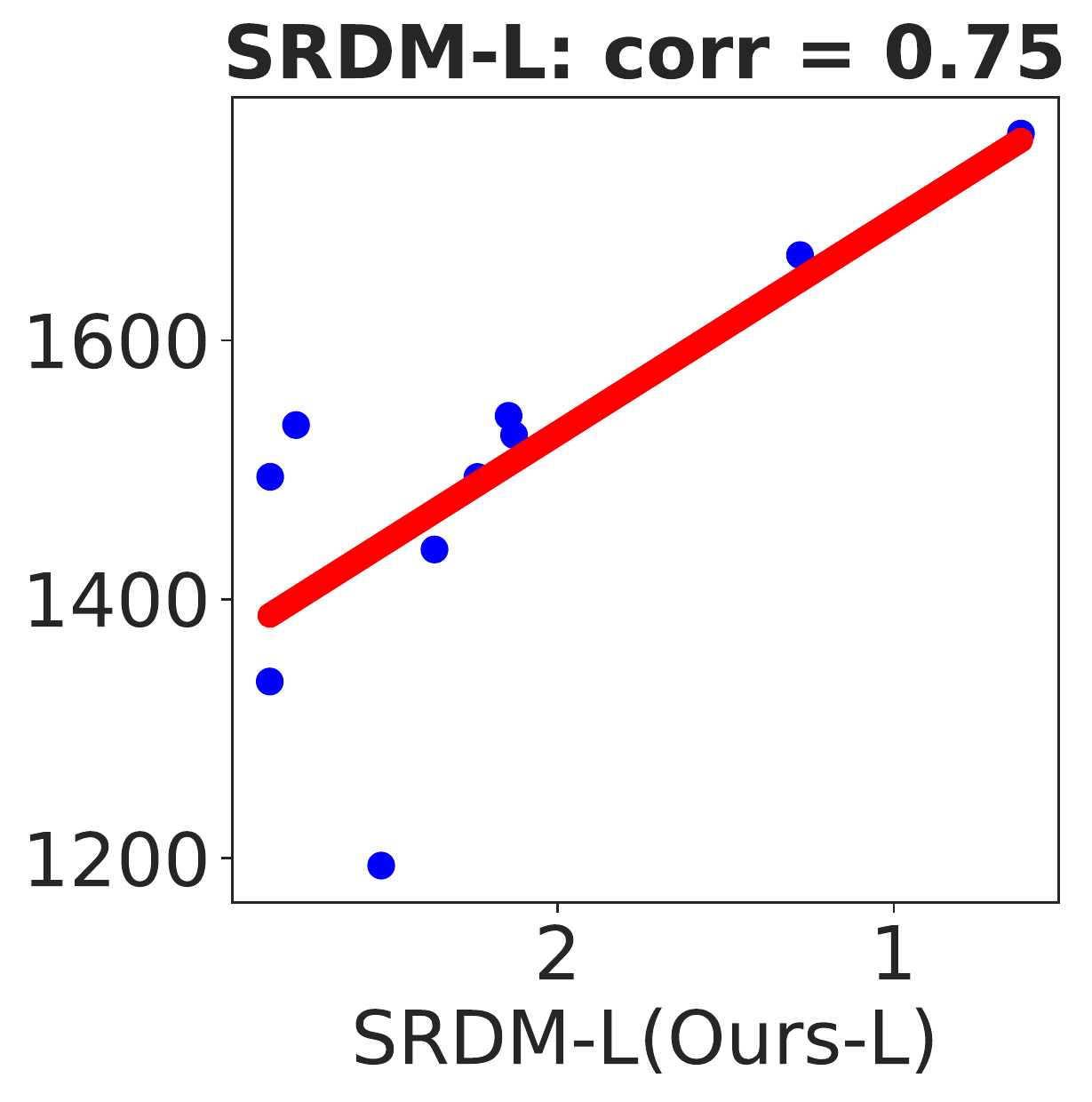} \\
      \hspace{-3mm}
    \includegraphics[width=0.43\linewidth]{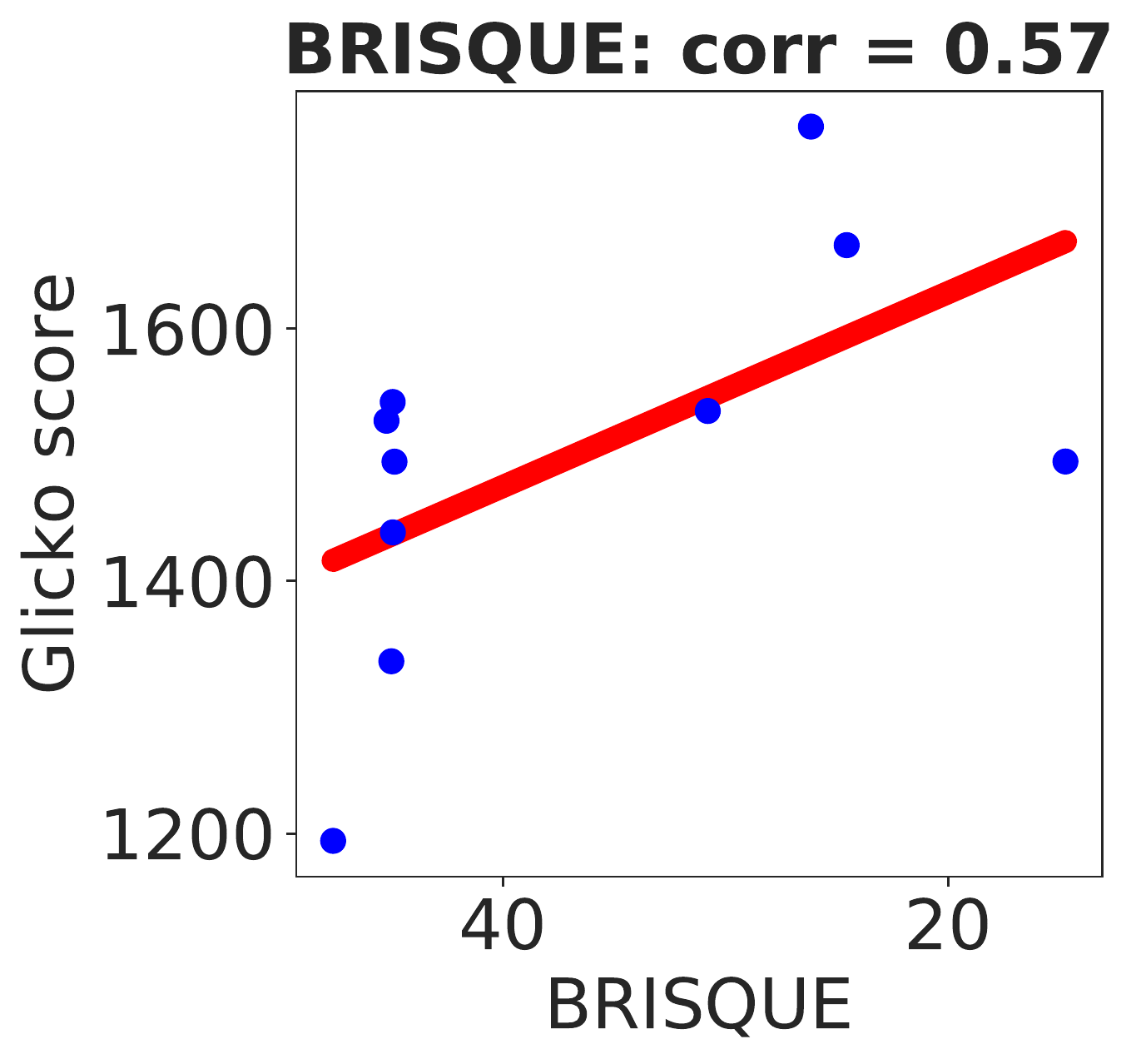} &
    \includegraphics[width=0.4\linewidth]{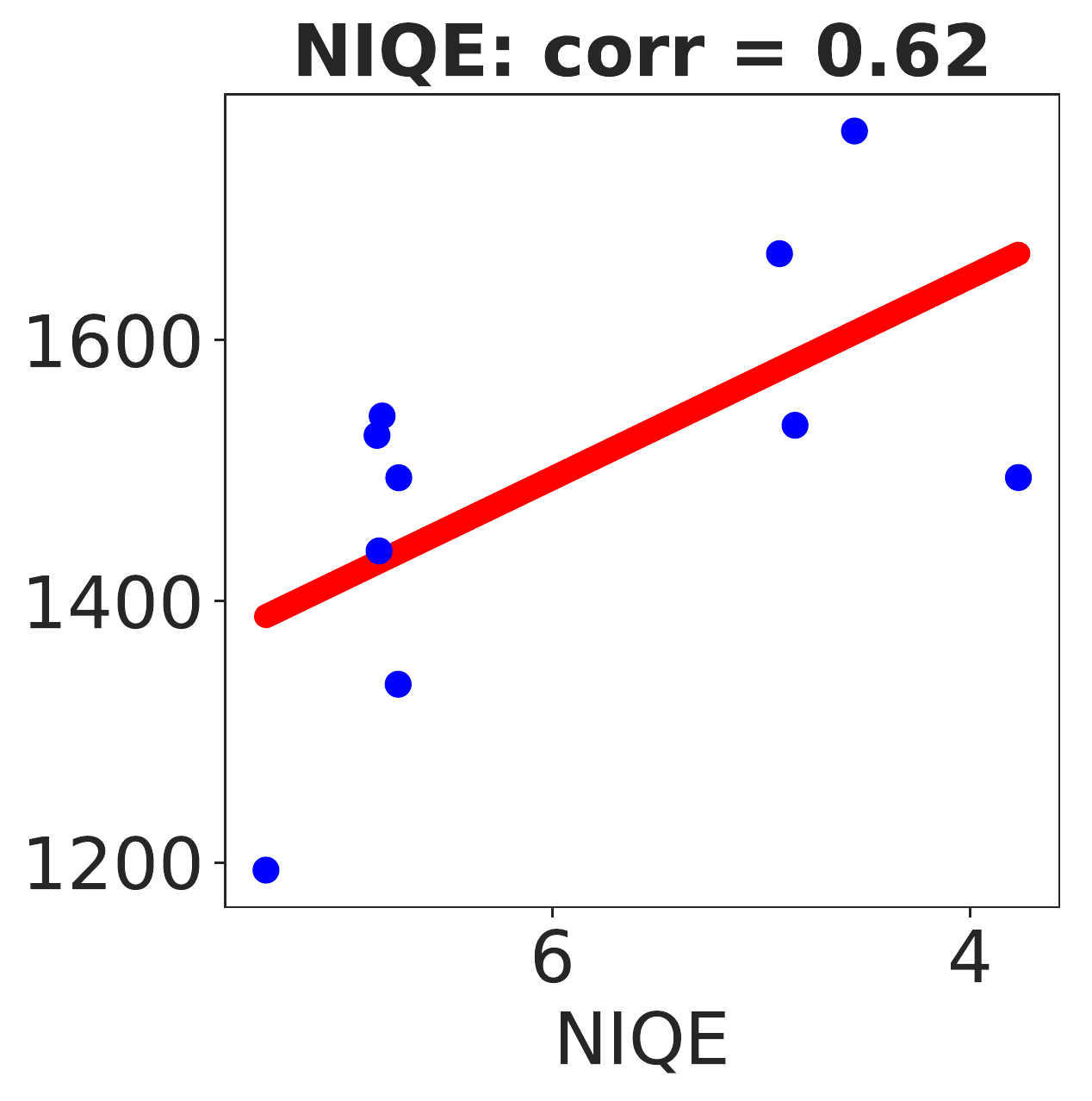} & 
      \includegraphics[width=0.4\linewidth]{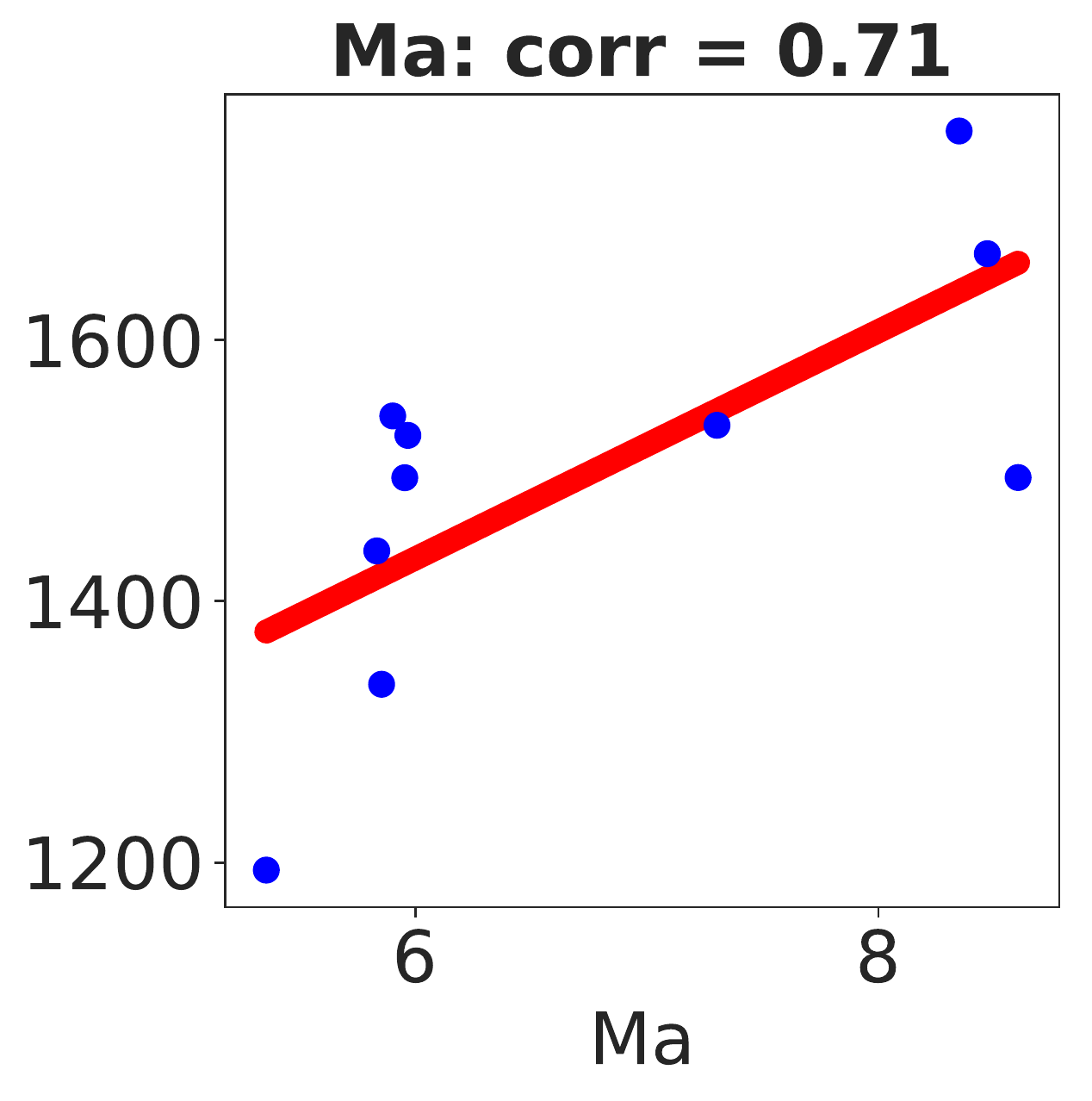}& 
    \includegraphics[width=0.4\linewidth]{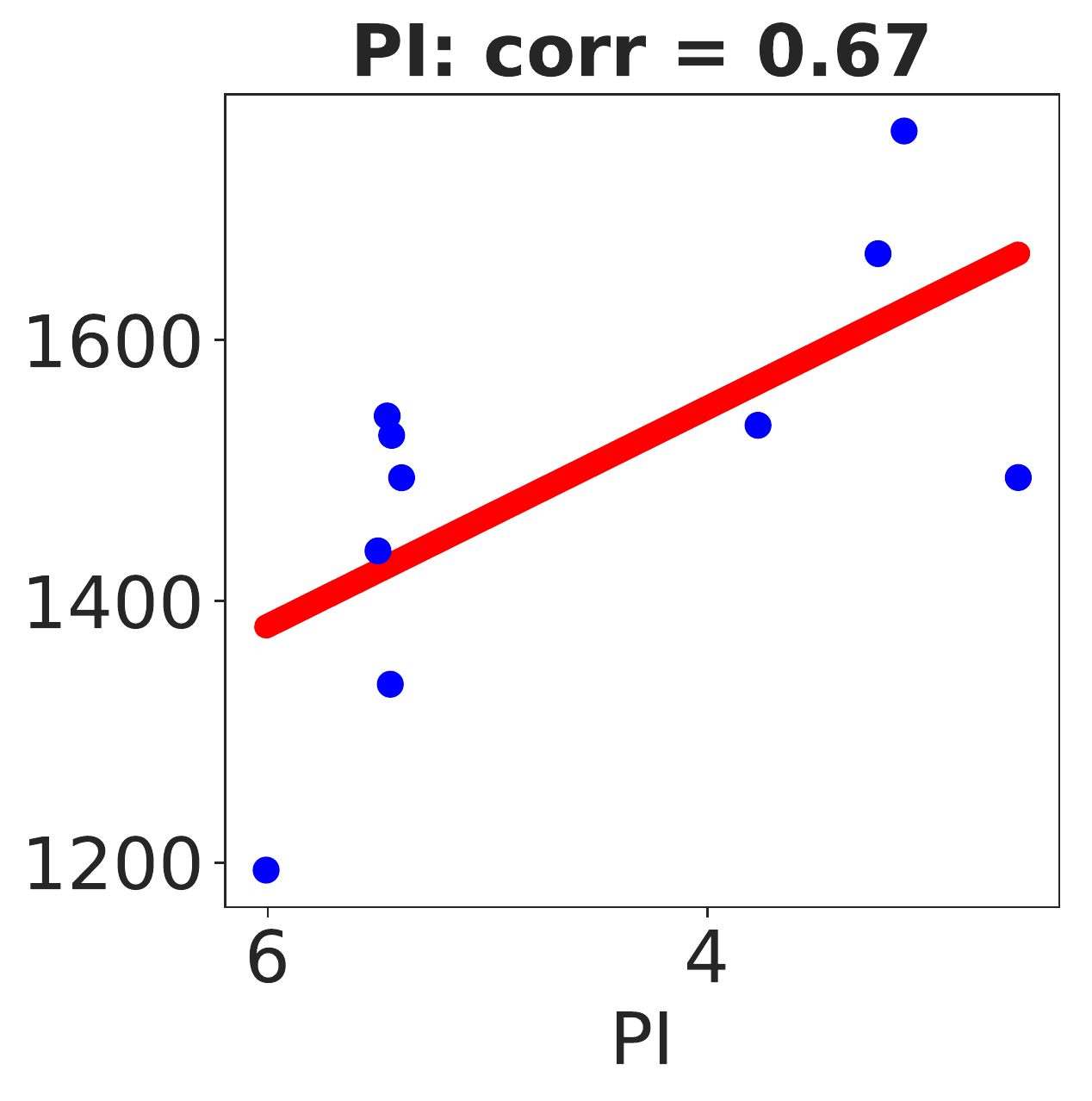} &
    \includegraphics[width=0.4\linewidth]{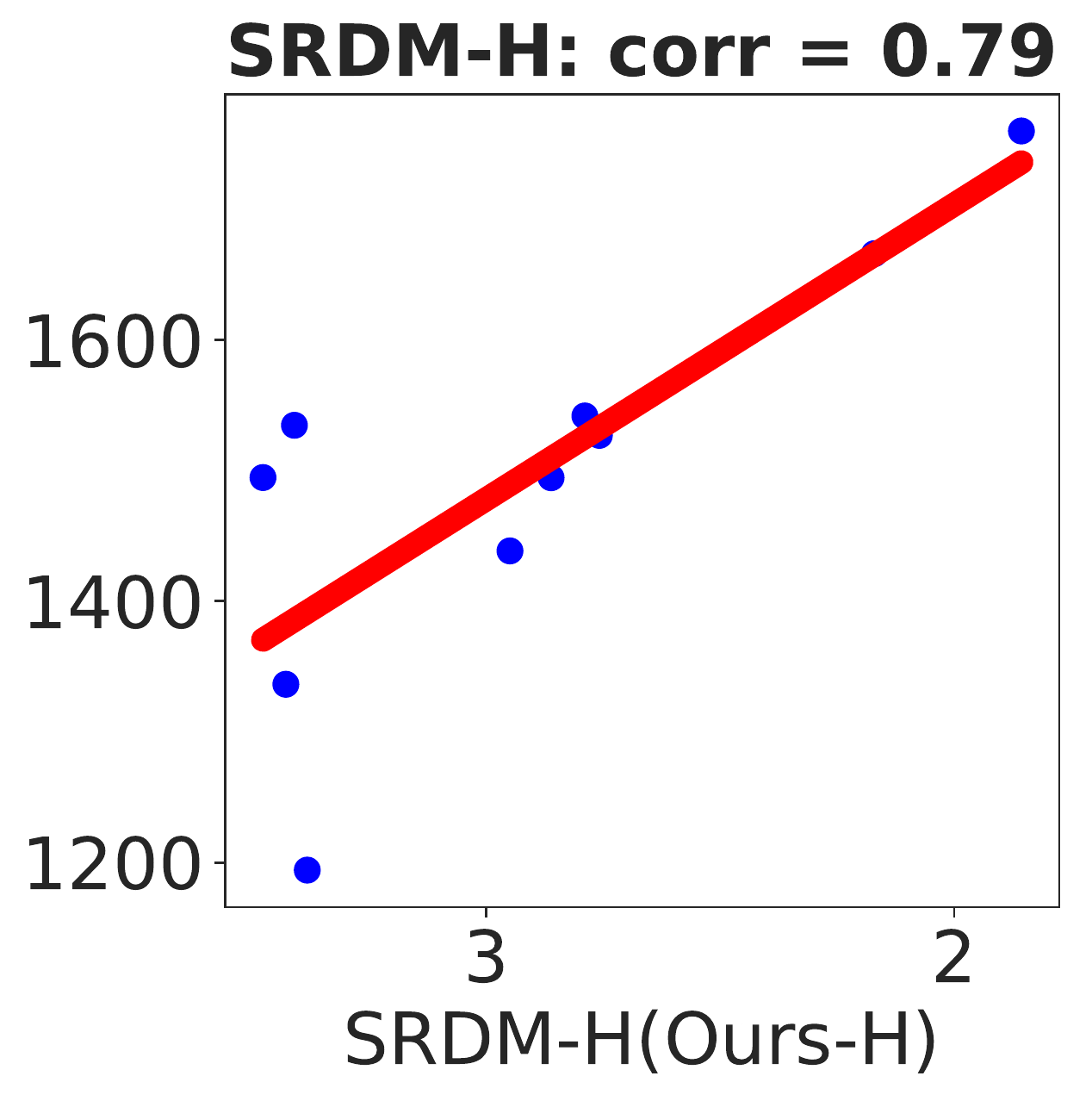}\\
    \end{tabular}
  \end{adjustbox}
  \caption{\textbf{Score correlation between SR evaluation metrics and the Glicko scores of the human subject study. } The figures show the Glicko scores of 9 state-of-the-art SR methods (y-axis) with respect to the mean scores using different SR evaluation metrics (x-axis), \ie, PSNR, SSIM~\cite{wang2004image}, IFC~\cite{sheikh2005information}, LPIPS~\cite{zhang2018unreasonable}, BRISQUE\cite{mittal2012no}, NIQE~\cite{mittal2012making}, Ma~\cite{ma2017learning}, PI~\cite{blau20182018} and the proposed metric. 
  We provide the results of the proposed metrics with two different grouping approaches, grouping on the original LR patch space (SRDM-H) and grouping on the 1-D projected space (SRDM-L).
  The proposed metric correlates well with the Glicko scores, and achieves similar performance compared to the best-performing deep-learning-based LPIPS.}
  \label{fig:correlation_figure}
  \vspace{-3mm}
\end{figure*}

\begin{figure*}[htb]
  \centering
  \Large
  \begin{adjustbox}{width=\linewidth}
    \begin{tabular}{ccccc}
    \hspace{-3mm}
    \includegraphics[width=35mm]{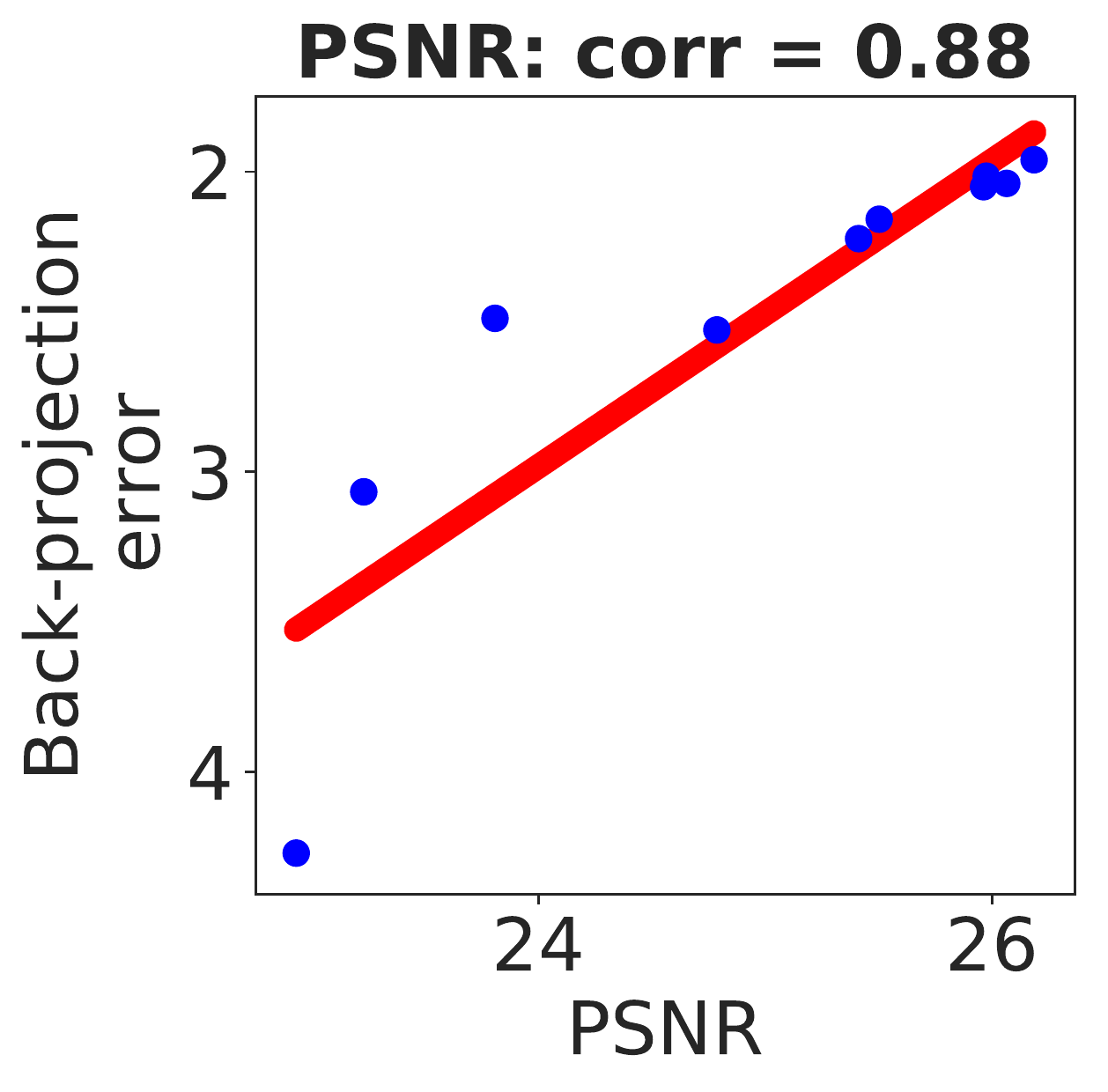} & \hspace{2mm}
\includegraphics[width=30.5mm]{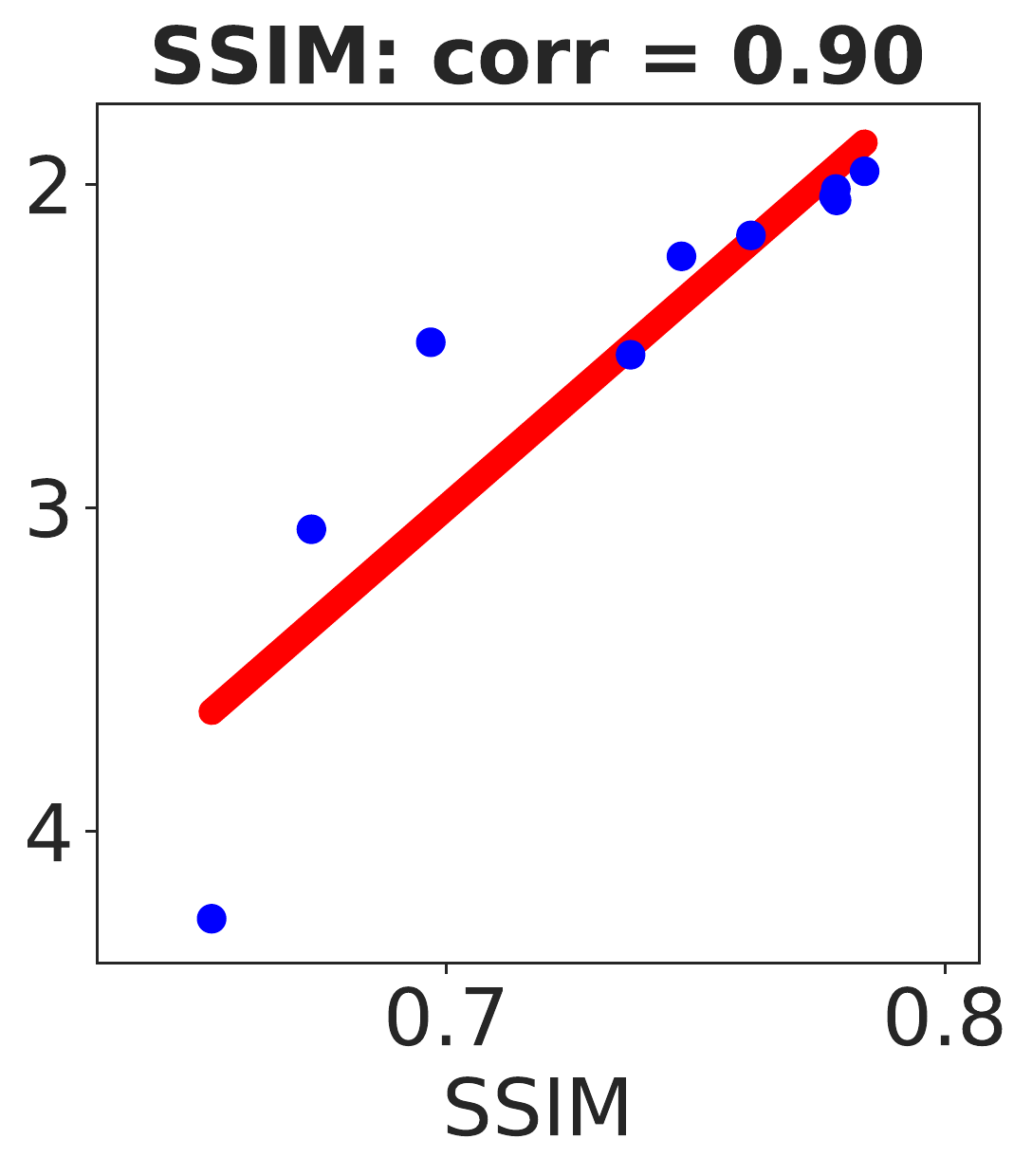} & \hspace{2mm}
    \includegraphics[width=31.5mm]{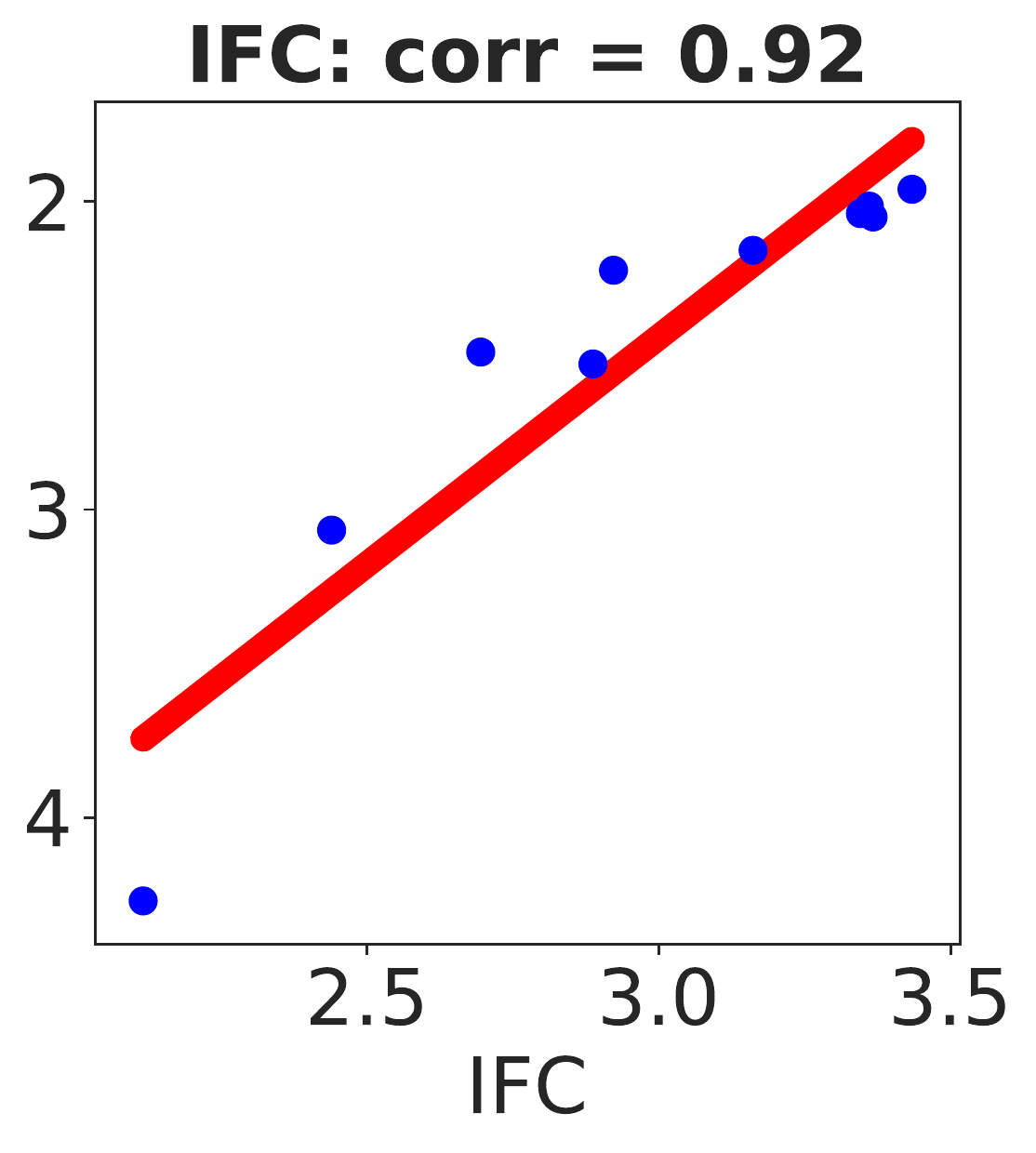} &\hspace{2mm}
    \includegraphics[width=29.5mm]{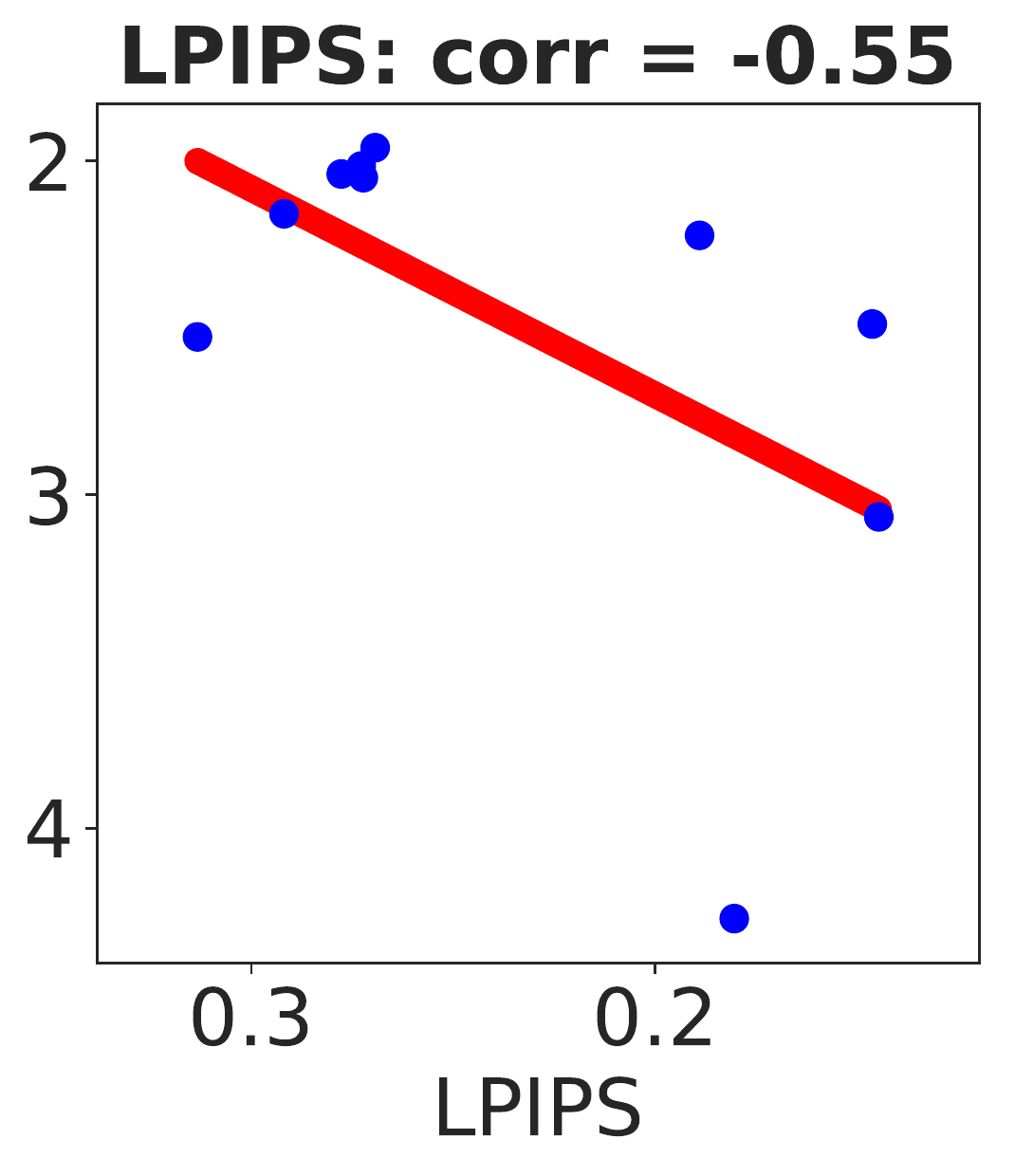} &\hspace{2mm}
  \includegraphics[width=29.5mm]{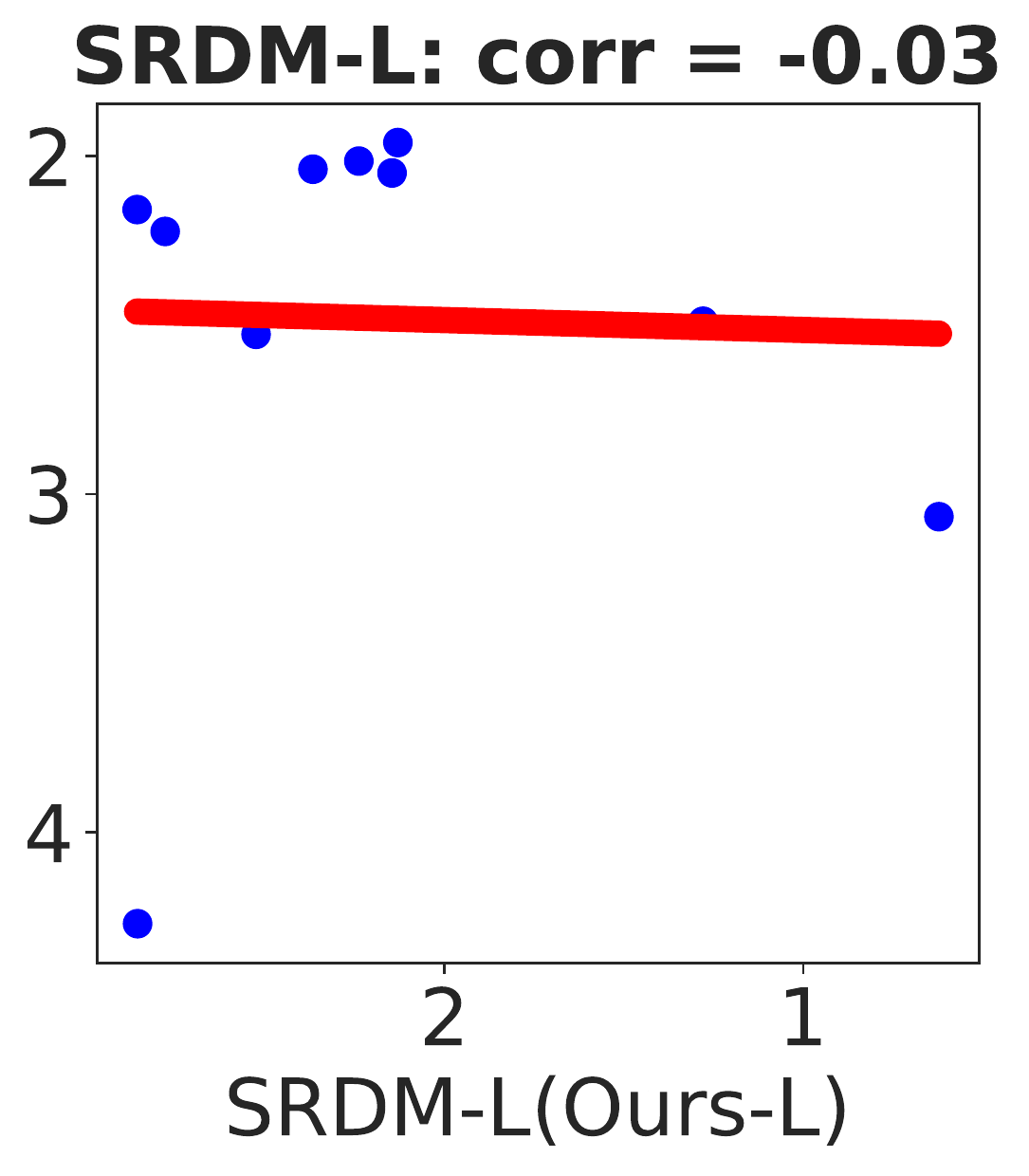} \\
    \hspace{-3mm}
    \includegraphics[width=37mm]{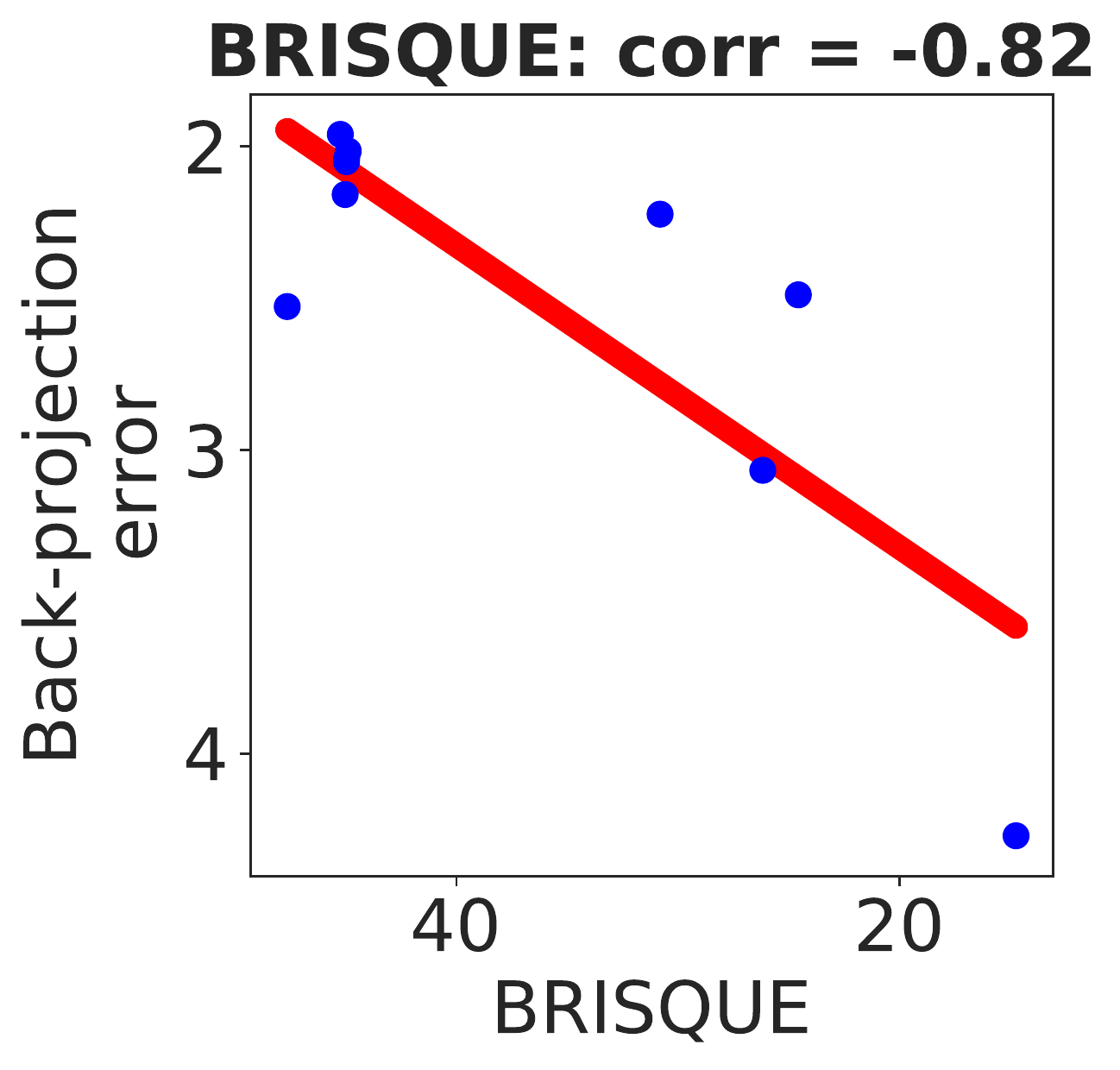} &\hspace{2mm}
    \includegraphics[width=30mm]{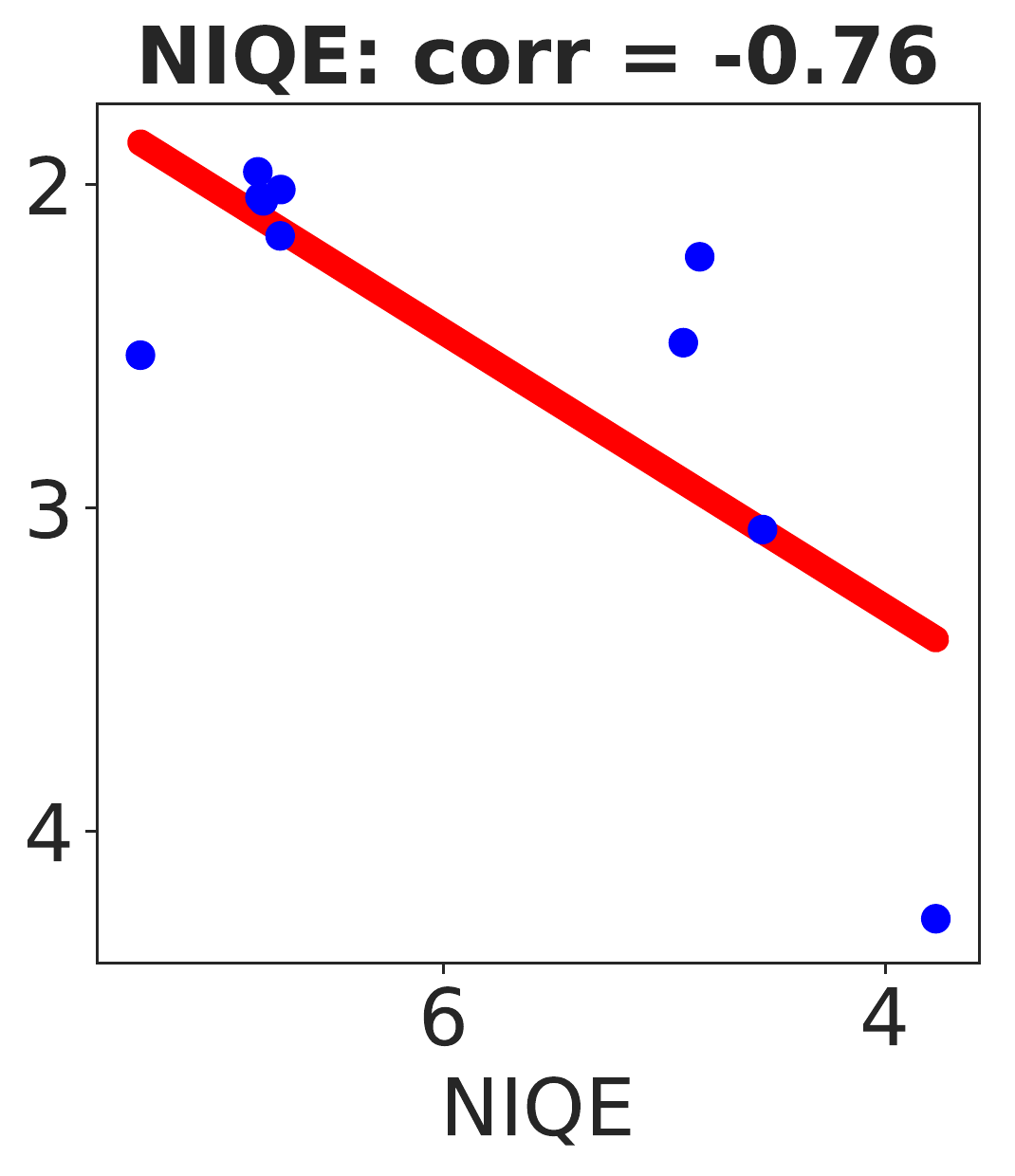} & \hspace{2mm}
     \includegraphics[width=30mm]{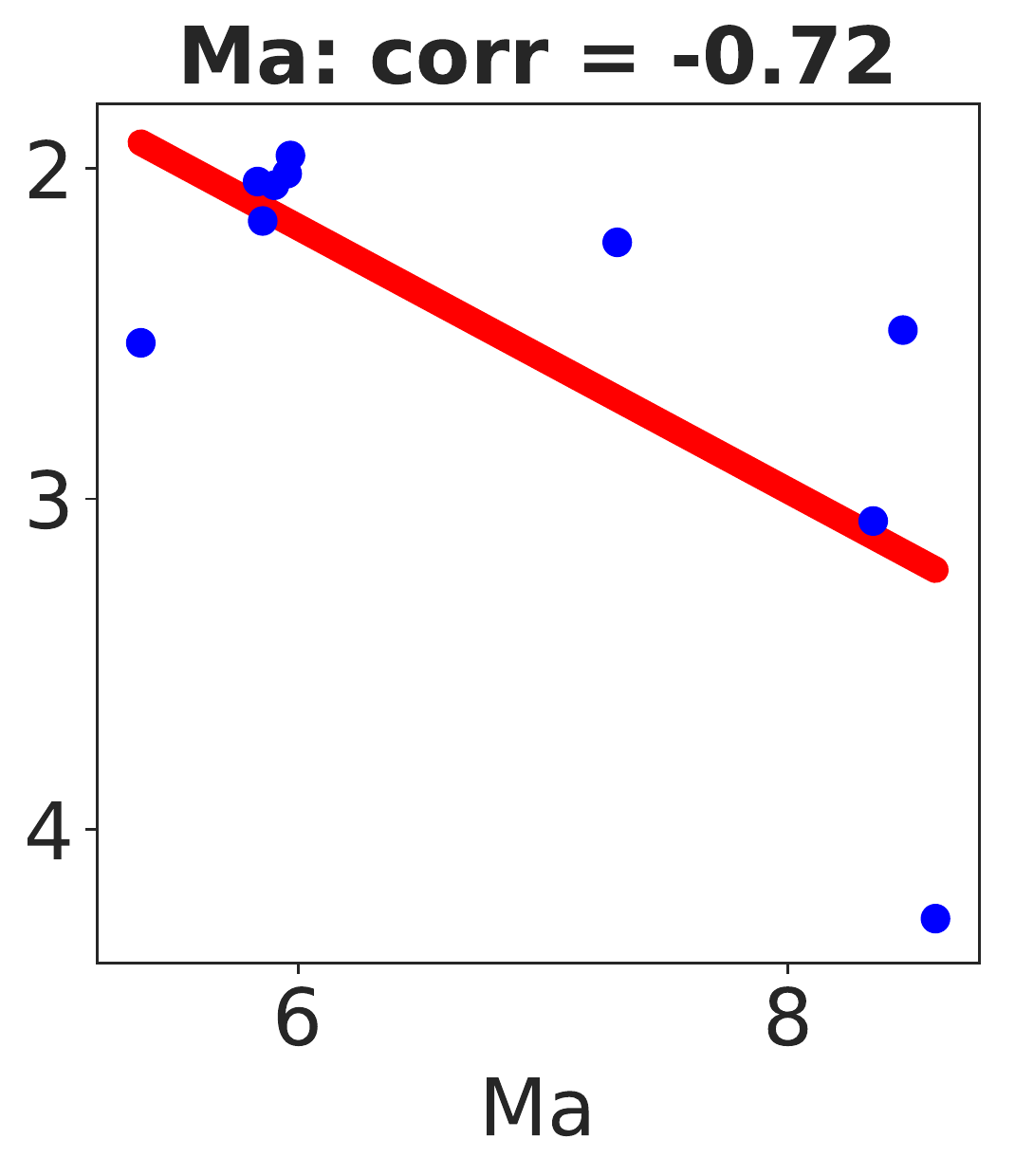}  &\hspace{2mm}
    \includegraphics[width=30mm]{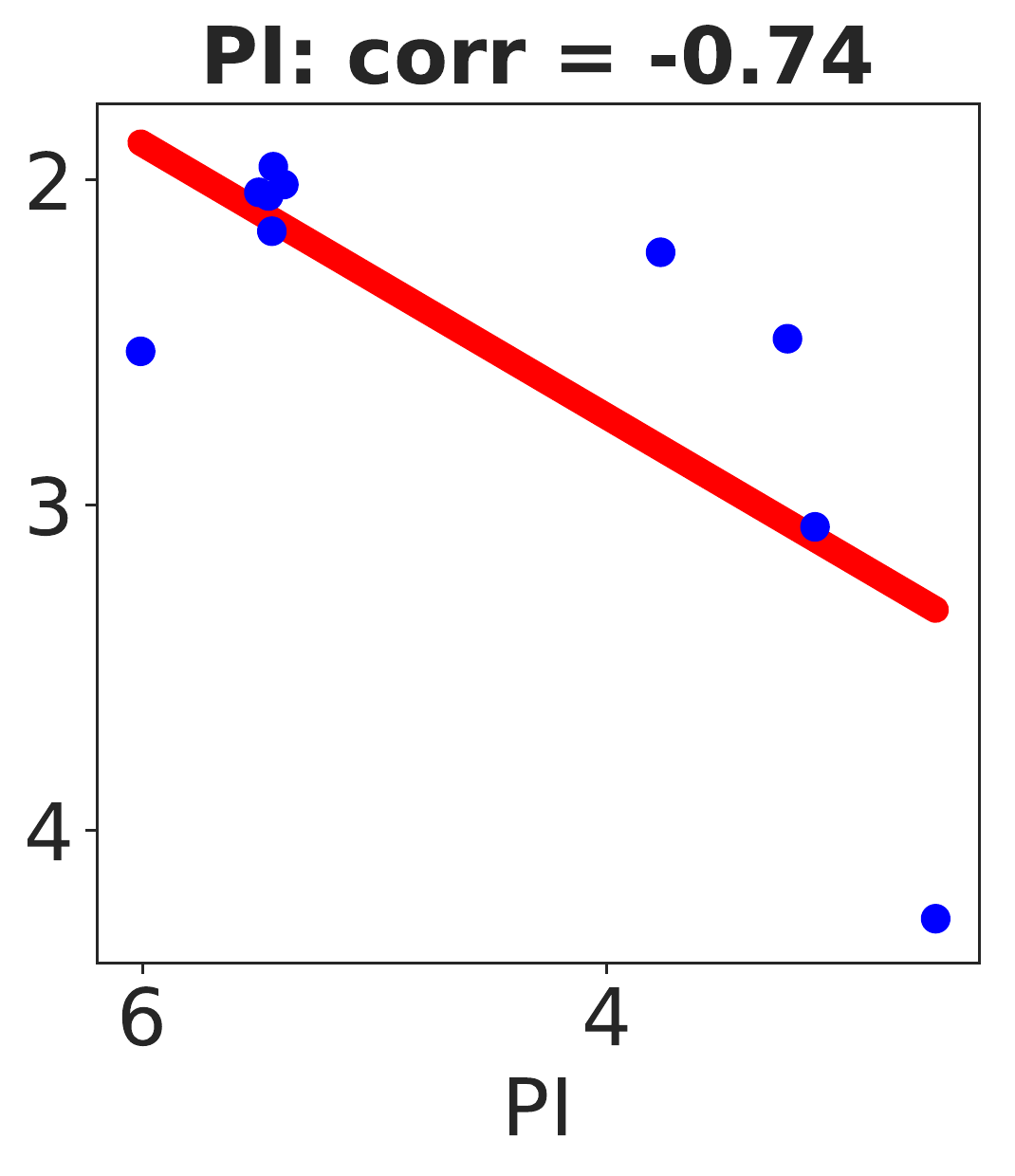} &\hspace{2mm}
    \includegraphics[width=30mm]{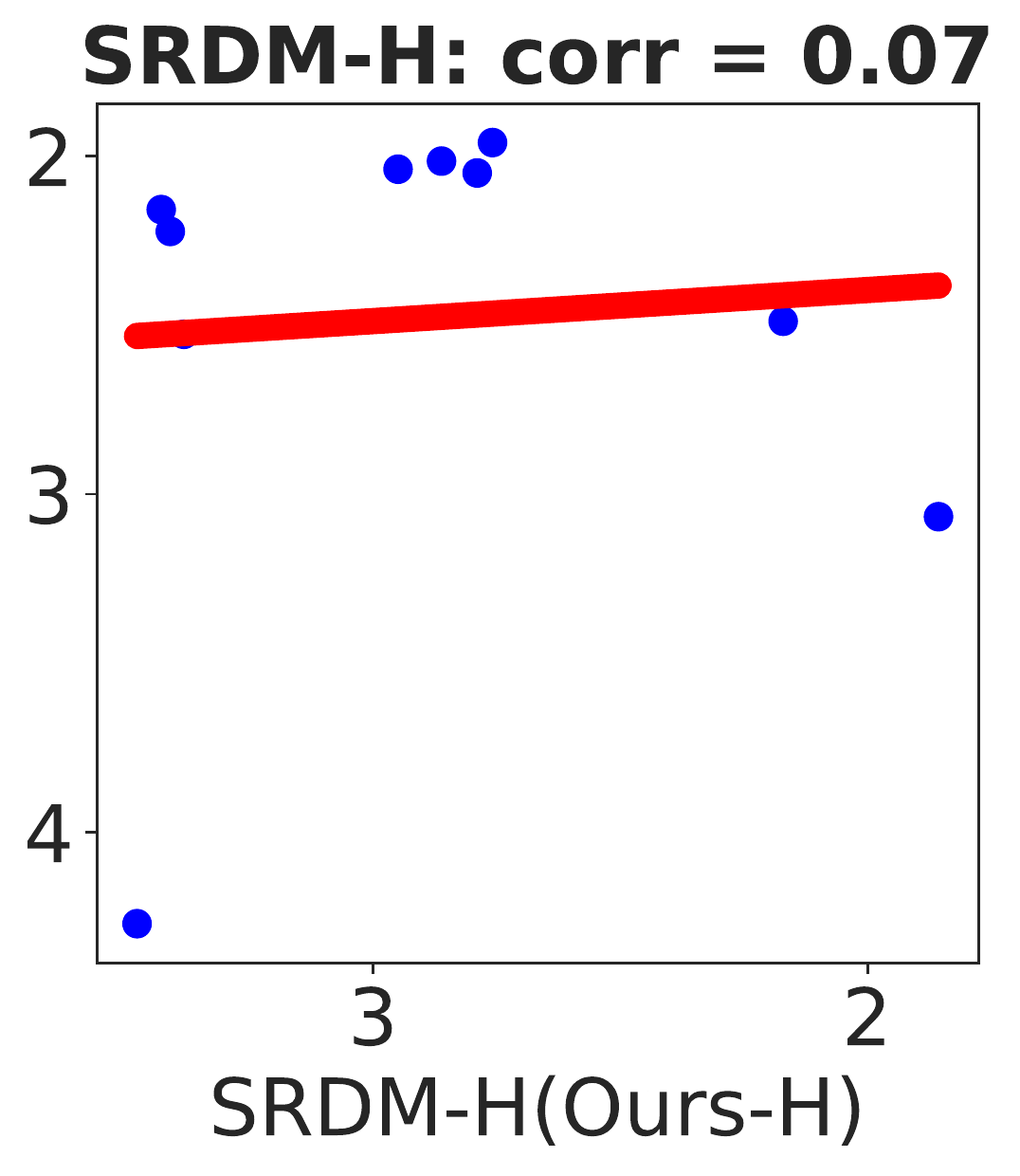}
    \\
    \end{tabular}
    \vspace{-2mm}
  \end{adjustbox}
  \caption{\textbf{Score correlation between SR evaluation metrics and the back-projection error.} The back-projection error is calculated using the RMSE(Root Mean Square Error) between the downsampled SR image and the input LR image. The figures show the back-projection error of 9 state-of-the-art SR methods (y-axis) with respect to the mean scores using different SR evaluation metrics (x-axis), \ie, PSNR, SSIM~\cite{wang2004image}, IFC~\cite{sheikh2005information}, LPIPS~\cite{zhang2018unreasonable}, BRISQUE~\cite{mittal2012no}, NIQE~\cite{mittal2012making}, Ma~\cite{ma2017learning}, PI~\cite{blau20182018}. 
  We provide the results of the proposed metrics with two different grouping approaches, grouping on the original LR patch space (Ours-H) and grouping on the 1-D projected space (Ours-L).
  The proposed metric is less correlated with the back-projection fidelity compared to PSNR, SSIM and IFC, however more correlated compared to LPIPS and the no-reference metrics.}
  \label{fig:correlation_figure_2}
  \vspace{-3mm}
\end{figure*}

To evaluate the quality of the generated HR patches, one naive method is to measure the distance between the distribution of the generated $Y$ and the distribution of the sampled ground truth $Y^*$, denoted as $D(Y, Y^*)$~\cite{blau2018perception}.
%
The distance is proper for evaluating the general distribution of $Y$, but not proper when $Y$ is generated from a conditional distribution.
As a toy example, we suppose $(x_1, y_1^*)$ and $(x_2, y_2^*)$ are the ground-truth LR-HR pairs, and $y_1$ and $y_2$ are the generated HR patches from $x_1$ and $x_2$, respectively. 
If $y_1 = y_2^*$ and $y_2 = y_1^*$, then $D(Y ,Y^*) = D( \{y_1, y_2\}, \{y_1^*, y_2^*\} )=  0$. 
However, the generated patches $y_1$ and $y_2$ are inaccurate for both $x_1$ and $x_2$, since they are swapped. 
A better metric in this case is the sum of the distance for HR patches of $x_1$ and the distance for HR patches of $x_2$ as
$D(\{y_1\},\{y_1^*\}) + D(\{y_2\},\{y_2^*\})$, which is non-zero. 
%

In general, an ideal metric would measure the quality of generated samples corresponding to each LR counter-part $x$ separately, and then aggregate them. 
%
Specifically, for any given $X = x$, we calculate the per-LR-patch distance by
$$d(p_{Y|X=x}, p_{Y^*|X=x}),$$
where the distance $d$ can be any distribution distance. 

Then the metric $D_0$ for the general quality of $Y$ can be defined as the expectation of per-LR-patch distance, 
\begin{equation}
\label{eq:d_instance} 
D_0((X,Y), (X,Y^*)) = \mathbb{E}_{ x \sim p_X} [d(p_{Y|X=x}, p_{Y^*|X=x})] . 
\end{equation}
Here we use $(X,Y)$ to represent the joint space of the LR-HR correspondences.
To measure the metric, it is required to get a good estimation for each $p_{Y^* |X = x}$, which needs a reasonable number of ground-truth samples $ y^* $ for each $X = x$. 
However, this is a difficult task in practice, as only a few ground-truth HR images are available for a single LR image. 
In most cases, we only obtain one sample drawn from $p_{ Y^* | X = x}$. 

To address the issue of insufficient samples, we propose a metric that deals with a group of $x$ at a time. 
Specifically, we split the LR patch space $\mathcal{X}$ into a set of non-overlapping groups $\mathcal{G} = \{G_i\}_{i=1}^{|\mathcal{G}|}$, where $G_i$ is a group consisting of a certain number of LR patches.
When the number of LR patches in a group $G$ is sufficiently large, it would be easy to estimate the distribution of HR patches corresponding to a group $G$, $p_{Y|X \in G}$.
%
%
Then the proposed metric can be defined as
\begin{equation}
\label{eq: first metric} 
D((X,Y), (X,Y^*)) = \frac{1}{|\mathcal{G}| } \sum_{ G \in \mathcal{G} } d( p_{Y|X \in G }, p_{Y^* | X \in G } ). 
\end{equation}
Compared with the metric~\eqref{eq:d_instance} by dealing with each $x$ separately (single-instance metric) and the naive metric by mixing all $x$ samples (``batch'' metric), the proposed metric can be viewed as an intermediate metric (``mini-batch'' metric).

The distribution for the HR image patch $Y$/$Y^*$ of size $h \times h$, is $256^{h \times h}$, if considering the 8-bit grayscale image.
It is impractical to compute the high-dimensional distribution for the distribution distance, even when $h$ is small. 
If $h=1$, it would be relatively easy to estimate $p_{Y| X \in G} $ and $ p_{Y^*|X \in G}$ with a sufficient number of samples.
%
%
Therefore, to resolve the issue, we project the HR patches $Y$ and $Y^* $ into 1-dimensional space, and collect the distribution in the 1-dimensional space.
%
That is, we consider the distribution $p_{w^T Y|X\in G}$ and $p_{w^T Y^*|X \in G}$, with a projection vector $w$ of the same size as $Y$. 
Then the metric can be written as,
\begin{equation}
\label{eq: second metric} 
D_w((X,Y), (X,Y^*)) 
  = \frac{1}{|\mathcal{G}|} \sum_{G \in \mathcal{G}} d( p_{w^T Y|X \in G}, p_{w^T Y^*|X \in G}). 
\end{equation}
%
%
%
We note that $p_Y$ represents a high-dimensional distribution while $p_{w^T Y}$ represents a low-dimensional distribution.
However, when considering the distribution distance, $d( p_{w^T Y|X \in G}, p_{w^T Y^*|X \in G})$ provides a valid approximation of $d(p_{Y|X \in G}, p_{Y^*|X \in G})$. 

\subsection{Implementation Details}
\label{subsec:implementation_details}
In the following, we describe in detail the design choices for the proposed metric $D_w$, including the grouping method, the projection vector, and the distribution distance $d$. 



\noindent\textbf{Grouping on LR patch $X$.} \
%
%
%
To resolve the issue of insufficient samples for computing the $Y$ distribution of a certain LR patch $X$, we split the LR patch space into groups $\mathcal{G}$ and calculate the $Y$ distribution for a group of LR patches. 
To make the metric $D$ a reasonable approximation to the ideal metric $D_0$, it is needed to enforce the group-based distribution $p_{Y|X \in G}$ to be similar to $\sum_{x \in G} p_{Y|X=x}$.
This can be achieved if the LR patches in each group are similar or correlated.
Therefore, an intuitive grouping strategy is the clustering directly on the LR patch space. 
When the total number of samples is huge, the clustering on the high-dimensional space becomes computational expensive. 
Therefore, it is suitable for one-time evaluation of SR results, but is less practical for tasks with limited resources, \eg, network training using the metric. 
In this work, we propose the metric with two different grouping methods for different purposes, one (referred as SRDM-H) with the clustering on the LR patch space and another (referred as SRDM-L) with the clustering on the projected 1-dimensional space of the LR patch, using the first principal component of LR patches.
We adopt the K-means method based on the Euclidean distance for clustering. 
The experiments in \secref{subsec:design_choice} show that the two approaches achieve similar performance, while SRDM-H is slightly better.

The number of groups $N_g$ is an important parameter, and is inversely correlated to the number of samples per group when the total number of samples is fixed. 
In fact, setting the number of groups either too large or too small is not optimal.
If the number of groups is too large, it is hard to get sufficient samples for calculating the distribution $p_{Y|X \in G}$ and $p_{Y^*|X \in G}$.
If the number of groups is too small, it indicates a large variety of LR patches in a group.
When the variety of LR patches in a group $G$ is close to that of all the LR patches, the distribution of the corresponding HR patches for $G$ would be also close to that for all the LR patches, which is the uniform distribution in the HR patch space.
In this case, it would be hard to distinguish $p_{Y|X \in G}$ and $p_{Y^*|X \in G}$, and therefore leads to less meaningful distribution distance.
In \secref{subsec:design_choice}, we will provide the empirical understanding on how to set the parameter based on the experiments on the number of groups.

\noindent\textbf{Projection on HR patch $Y$.} \
%
%
%
To calculate the $Y$ distribution in practice, we collect the distribution on the subspace of $Y$ using a projection vector $w$. 
And the distribution dimension after the projection would be $256$, if considering the 8-bit grayscale image.
There are many choices of the projection vector $w$. 
As an idea metric for SR is to represent the pixel-wise accuracy, using a projection vector that selects a certain pixel is more reasonable compared than that averages many pixels. 
Moreover, considering that the neighboring pixels in the LR patch could also help to determine the SR of the center pixel, we propose to use the projection vector that selects a pixel in the HR patch corresponding to the center pixel of the LR patch. 
There are $s \times s$ HR pixels corresponding to the center pixel of the LR patch, where $s$ is the factor of the SR. 
In fact, we find that the evaluation metric is robust with respect to the pixel selection within the $s \times s$ region, as shown in \secref{sec:exp}. 
Therefore, we use the center pixel of the region (coordinate $(\frac{s-1}{2},\frac{s-1}{2})$ if $s$ is even) as the default in the following experiments.

\noindent\textbf{Distribution distance $d$.} \
For the distribution distance $d$ in the metric, there are many choices, \eg, the total variation (TV) distance, the Killback-Leibler (KL) divergence, the Jensen-Shannon (JS) divergence, and the Wasserstein distance.
However, not all the distances are suitable for this problem. 
The TV distance measures the largest possible difference between two distribution, and does not give accurate estimation on the general difference.
The KL divergence and the JS divergence are commonly used distribution distances, but they both suffer from issues when the two distributions have disjoint support (\ie, the KL divergence would explode and the JS divergence would remain a constant).
The Wasserstein distance measures the cost of moving one distribution to the other distribution, and is commonly used for image comparison metrics~\cite{rubner2000earth}. 
Moreover, we will show that the metric using the Wasserstein distance is more robust in terms of other hyper-parameters, in \secref{sec:exp}.
Therefore, we propose to use the Wasserstein distance for $d$ as the default. 
\section{Experiment Results}
\label{sec:exp}
In this section, we conduct experiments to show the benefits of the proposed metric. 
We demonstrate that the proposed metric matches the perceptual quality of the HVS. 
%
We verify the robustness of the proposed metric with respect to the metric parameters.
Moreover, we show how to use the proposed metric as the loss for training SR networks. 
Without loss of generality, we conduct analysis and experiments on 4$\times$ SR case in this work.

\begin{table}[t]
{\scriptsize  
    \begin{minipage}[t]{0.44\linewidth}
        \begin{center}
        \renewcommand{\arraystretch}{1.1}
        \begin{tabular}[t]{ccc}
        \hline
        Method & Rating Score  & Uncertainty\\
        \hline
        SRResNet~\cite{ledig2017photo} &1336.408&64.796 \\
        SRGAN~\cite{ledig2017photo} &1494.593&62.901 \\
        Lapsrn~\cite{lai2018fast}& 1194.190  &69.350  \\
        RCAN~\cite{Zhang_2018_ECCV} & 1541.713   &63.197 \\
        EDSR~\cite{lim2017enhanced} &1494.451  &62.911  \\
        EPSR~\cite{vasu2018analyzing} &1534.584&63.280 \\
        ESRGAN(PSNR)~\cite{Wang_2018_ECCV_Workshops}&1526.869&62.257 \\
        ESRGAN(GAN)~\cite{Wang_2018_ECCV_Workshops} &1759.780&65.555 \\
        ProSR(PSNR)~\cite{wang2018fully} &1438.452&62.598 \\
        ProSR(GAN)~\cite{wang2018fully} &1665.900&64.605 \\
        \hline
        \end{tabular}
        \caption{\textbf{Glicko~\cite{glickman1995glicko} rating results on the human subject study.} The Glicko scores approximate the perceptual quality of SR results. High scores indicate high image perceptual quality.}
        \label{tab:survey_score}%
        \end{center}
    \end{minipage}
    \hfill
    \begin{minipage}[t]{0.48\linewidth}
        \begin{center}
        \renewcommand{\arraystretch}{1.4}
        \begin{tabular}[t]{lcccc}
        \hline
        Setting & PNSR & SSIM & Ma~\cite{ma2017learning} & LPIPS~\cite{zhang2018unreasonable}\\
        \hline
        RCAN~\cite{Zhang_2018_ECCV}   &  29.28& 0.8406& 4.746&0.2552 \\    
        $r$=25 $N_g$=120 & 28.74&0.8365&5.278&0.2521 \\
        $r$=19 $N_g$=120 & 28.77&0.8382&5.282&0.2490 \\
        $r$=13 $N_g$=120 & 28.44&0.8302&5.374&0.2549 \\
        \hline
        $r$=25 $N_g$=80 & 28.83&0.8391&5.164 &0.2534\\
        $r$=25 $N_g$=120 & 28.74&0.8365&5.278&0.2521 \\
        $r$=25 $N_g$=160 & 28.73&0.8368&5.290&0.2509 \\
        \hline
        \end{tabular}
        \caption{\textbf{Quantitative results of the finetuned RCAN~\cite{Zhang_2018_ECCV} models using the proposed metric.} We evaluate two parameter settings, LR patch size $r$ and number of groups $N_g$. All the model are finetuned from the pre-trained RCAN model (first row) using $\ell_1$ loss.}
        \label{tab:training_table}
        \end{center}
    \end{minipage}
}
\vspace{-4mm}
\end{table}
\subsection{Human Subject Study}
To validate the effectiveness of the proposed metric, we conduct a human subject study. 
%
%
Among numerous existing SR methods, we evaluate 10 representative state-of-the-art methods in the experiment, \ie  SRResNet~\cite{ledig2017photo}, SRGAN~\cite{ledig2017photo}, LapSRN~\cite{lai2018fast}, RCAN~\cite{Zhang_2018_ECCV}, EDSR~\cite{lim2017enhanced}, EPSR~\cite{vasu2018analyzing}, ESRGAN(PSNR)~\cite{Wang_2018_ECCV_Workshops}, ESRGAN(GAN)~\cite{Wang_2018_ECCV_Workshops}, ProSR(PSNR)~\cite{wang2018fully} and ProSR(GAN)~\cite{wang2018fully}. 
We include the methods trained with only the $\ell_1$/$\ell_2$ loss (\eg, SRResNet, EDSR, ESRGAN(PSNR) and ProSR(PSNR)) or with the adversarial loss (\eg, SRGAN, EPSR, ESRGAN(GAN) and ProSR(GAN)).

We use the commonly used DIV2K validation dataset~\cite{timofte2018ntire} to carry out the experiments. 
We apply the evaluated SR methods, using their provided source codes and trained models, on the 100 images of the DIV2K validation dataset and collect the estimated SR images. 
%

For the human subject study, we adopt the pairwise comparison approach that asks each human subject to choose a preferred image from a pair of estimated SR images. 
%
To help the human subject easily distinguish the difference between compared images, we select a 400$\times$400 image region for each DIV2K image that has the most diverged estimation results. 
We calculate the dispersion index (variance-to-mean ratio) among SR results of different methods at each pixel, and the image region with the maximum average dispersion is chosen.
We develop a web-based system to randomly select and present the comparison pair of 400$\times$400 image regions. 
We collect more than 1000 votes from 65 human subjects. 
There are in total ${10 \choose 2} = 45$ pairs of method comparisons, and each pair of methods receives more than 20 votes in average.
According to the user feedback, it is often quite easy for the
human subjects to choose the preferred SR results.

{\flushleft \bf Glicko rating results.}
%
To compute the ranking from paired comparison votes, we use the Glicko rating system~\cite{glickman1995glicko} that is commonly used to evaluate players in paired competitions~\cite{barber2012bayesian} with a consideration on the rating reliability.
The Glicko system estimates both rating $r$ and uncertainty $\sigma$, and the 
final rating is based on the range $(r-1.96\sigma, r+1.96\sigma)$ with a 95\% confidence. 
Therefore, the method with higher score and lower uncertainty will be ranked higher.
To remove the influence of sequence ordering, we feed in randomly shuffled vote results and repeat the process several times.

The Glicko rating scores and final rankings are shown in Table~\ref{tab:survey_score}. 
The Glicko scores approximate the perceptual quality of the SR results, and higher scores indicate higher image perceptual quality.
In general, the models trained with the adversarial loss obtain results of higher qualities, which also matches the observations in existing works~\cite{ledig2017photo,lai2018fast,Wang_2018_ECCV_Workshops}.

\subsection{Relation to Perceptual Quality} 
\label{subsec:RelationtoPQ}
In this section, we analyze the relation between the proposed metric and the human subject study results. 
For a thorough study, we provide the comparisons with 8 commonly used metrics for SR evaluation, \ie, PSNR, SSIM~\cite{wang2004image}, IFC~\cite{sheikh2005information}, LPIPS~\cite{zhang2018unreasonable}, BRISQUE\cite{mittal2012no}, NIQE~\cite{mittal2012making}, Ma~\cite{ma2017learning}, and PI~\cite{blau20182018}. 
We evaluate the SR results of the 10 SR methods using different metrics.
Similar to \cite{blau20182018}, we plot the figures of the Glicko scores of the SR methods, over the mean scores based on the compared metrics, in \figref{fig:correlation_figure}. 
The red lines in the figures are the least-squares regression lines.
Moreover, we report the Pearson correlation coefficient~\cite{pearson1895vii} with the Glicko scores.

As shown in \figref{fig:correlation_figure}, the full-reference metrics, PSNR, SSIM and IFC, are negatively correlate with the perceptual quality. 
The no-reference metrics, BRISQUE, NIQE, Ma and PI, have high correlation scores. 
The proposed metric is more correlated with the perceptual quality compared to other metrics, and is comparable with the deep learning based metric LPIPS.

\subsection{Relation to SR Fidelity} 
\label{subsec:RelationtoSR}
Other than the perceptual quality, we investigate the correlation between the proposed metric and the SR fidelity.
To represent the SR fidelity, we adopt the back-projection error that evaluates the RMSE between the downsampled SR image and the input LR image, \ie, $||g(Y) - X||_2$. 
%
We calculate the back-projection error on the SR images of different SR methods, and plot the correlation curves between the back-projection error and SR evaluation metrics in \figref{fig:correlation_figure_2}. 
The fidelity-based metrics, PSNR, SSIM and IFC, highly correlate with the back-projection error, while the perceptual quality based metrics are anti-correlated, as expected. 
Different from the metric LPIPS that does not maintain the fidelity, the proposed method has a better correlation with the fidelity, but less correlated compared with the fidelity-based metrics.
%
Therefore, the proposed metric provide a reasonable trade-off between the perceptual quality and the fidelity.

\begin{figure*}[htb]
\centering
\begin{adjustbox}{width=\linewidth}
\begin{tabular}{cccc}
\vspace{-1mm}
\hspace{-3mm}
\includegraphics[width=0.4\linewidth]{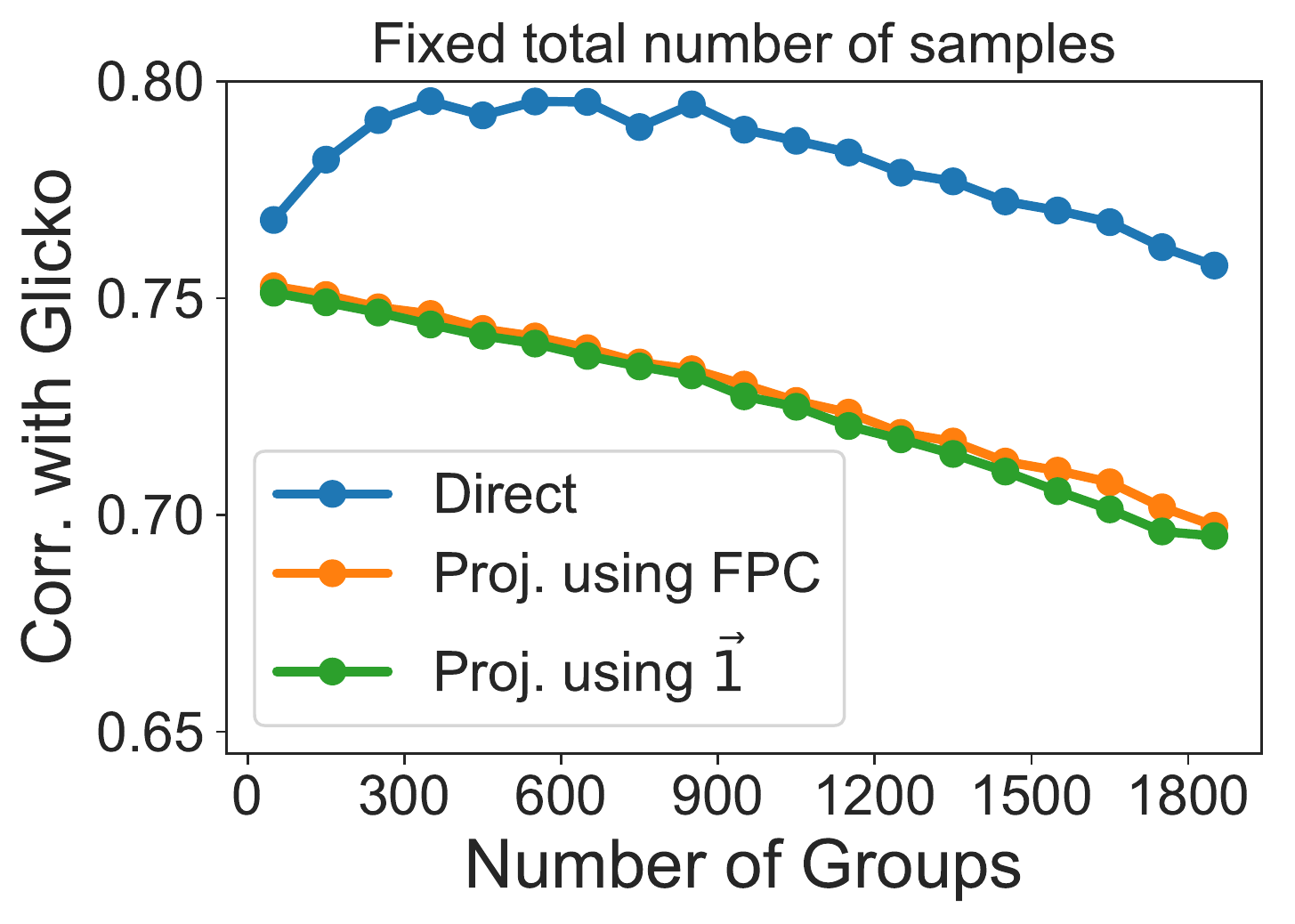} & 
\includegraphics[width=0.42\linewidth]{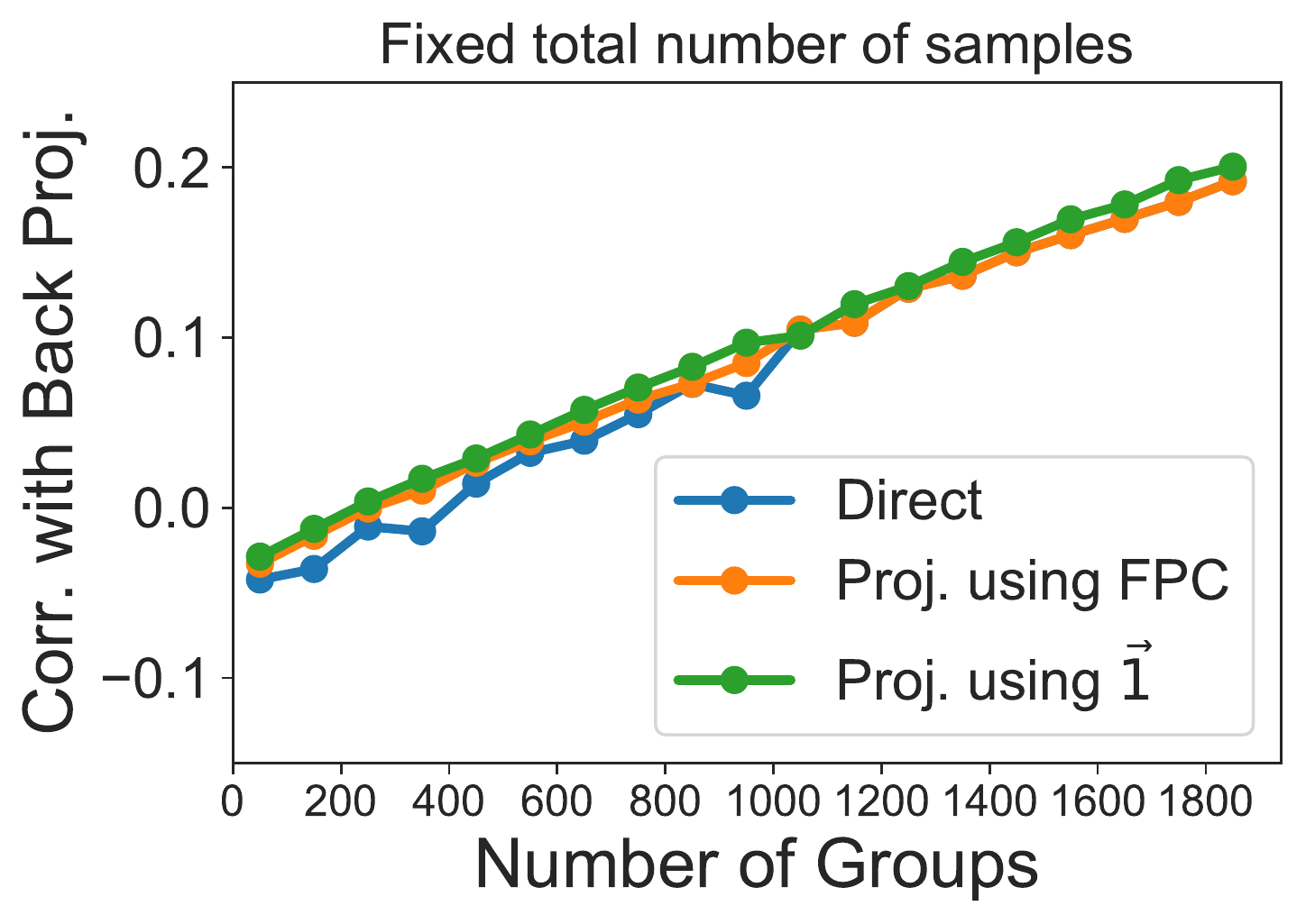} &
\includegraphics[width=0.4\linewidth]{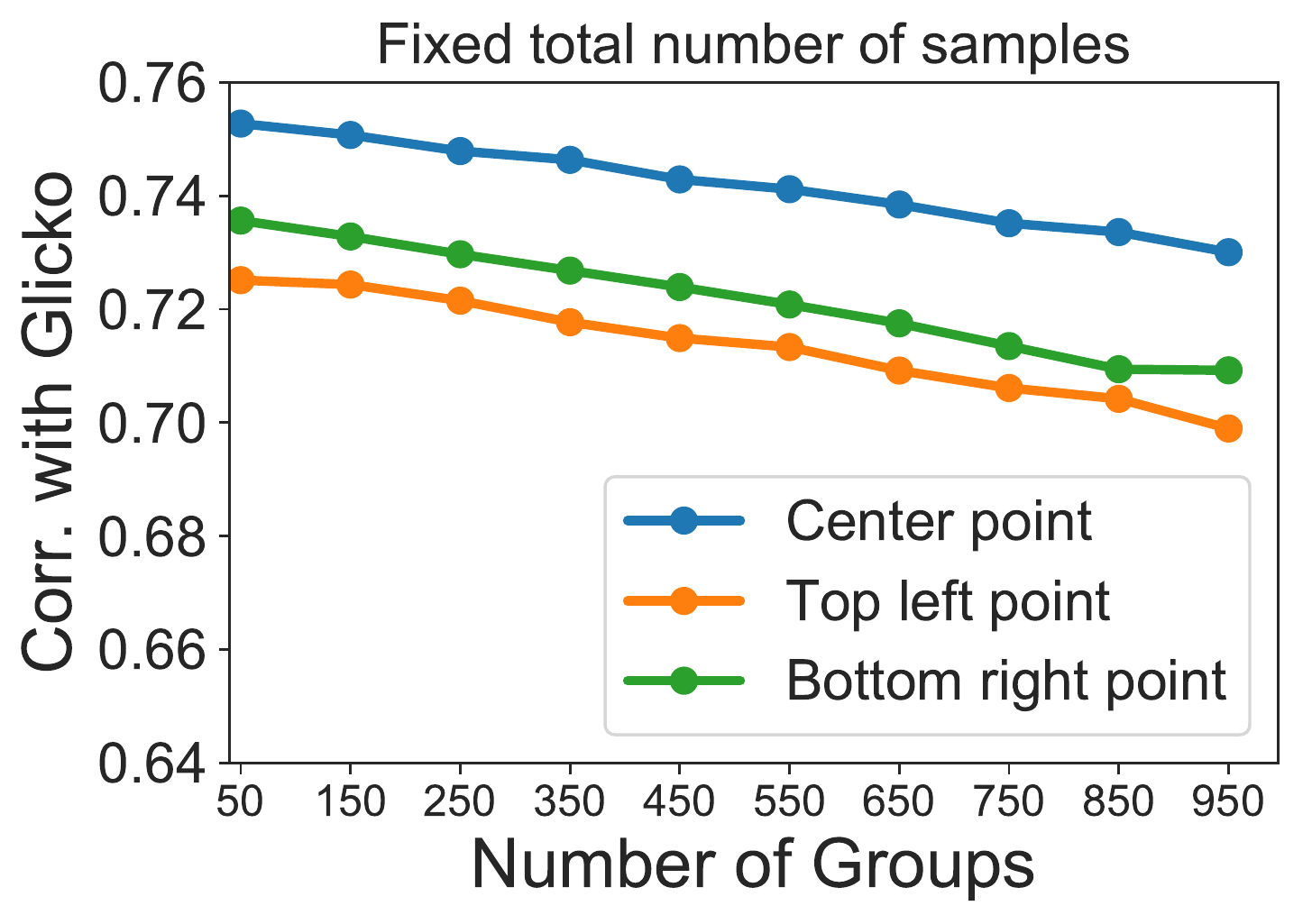} &
\includegraphics[width=0.4\linewidth]{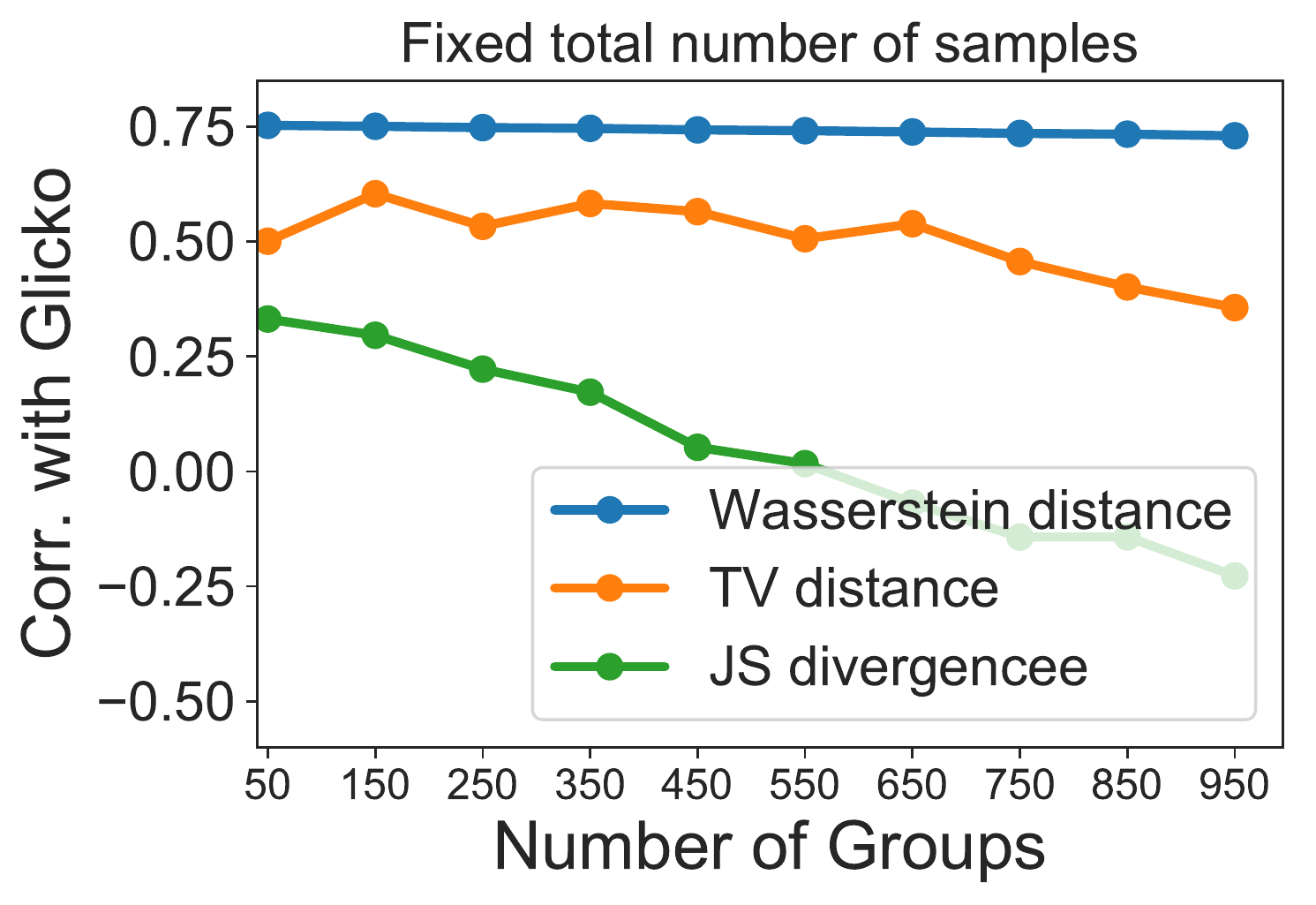} 
\\ \vspace{1mm}
\hspace{4mm} (a) Grouping method on $X$ & \hspace{8mm} (b) Grouping method on $X$ & \hspace{8mm} (c) Projection on $Y$ &\hspace{7mm}  (d) Distribution distance $d$
\\
\hspace{-3mm}
\includegraphics[width=0.41\linewidth]{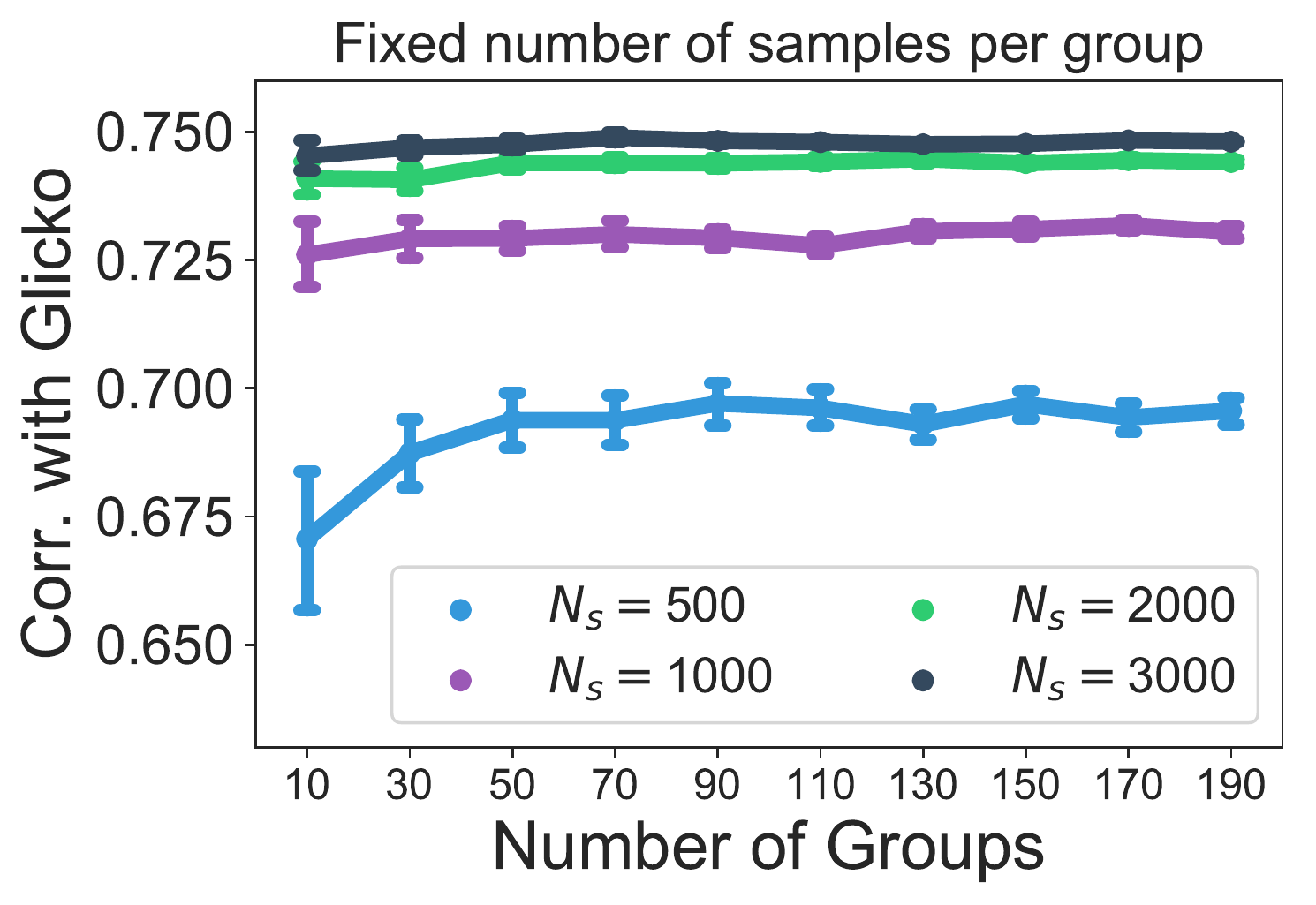} &
\includegraphics[width=0.4\linewidth]{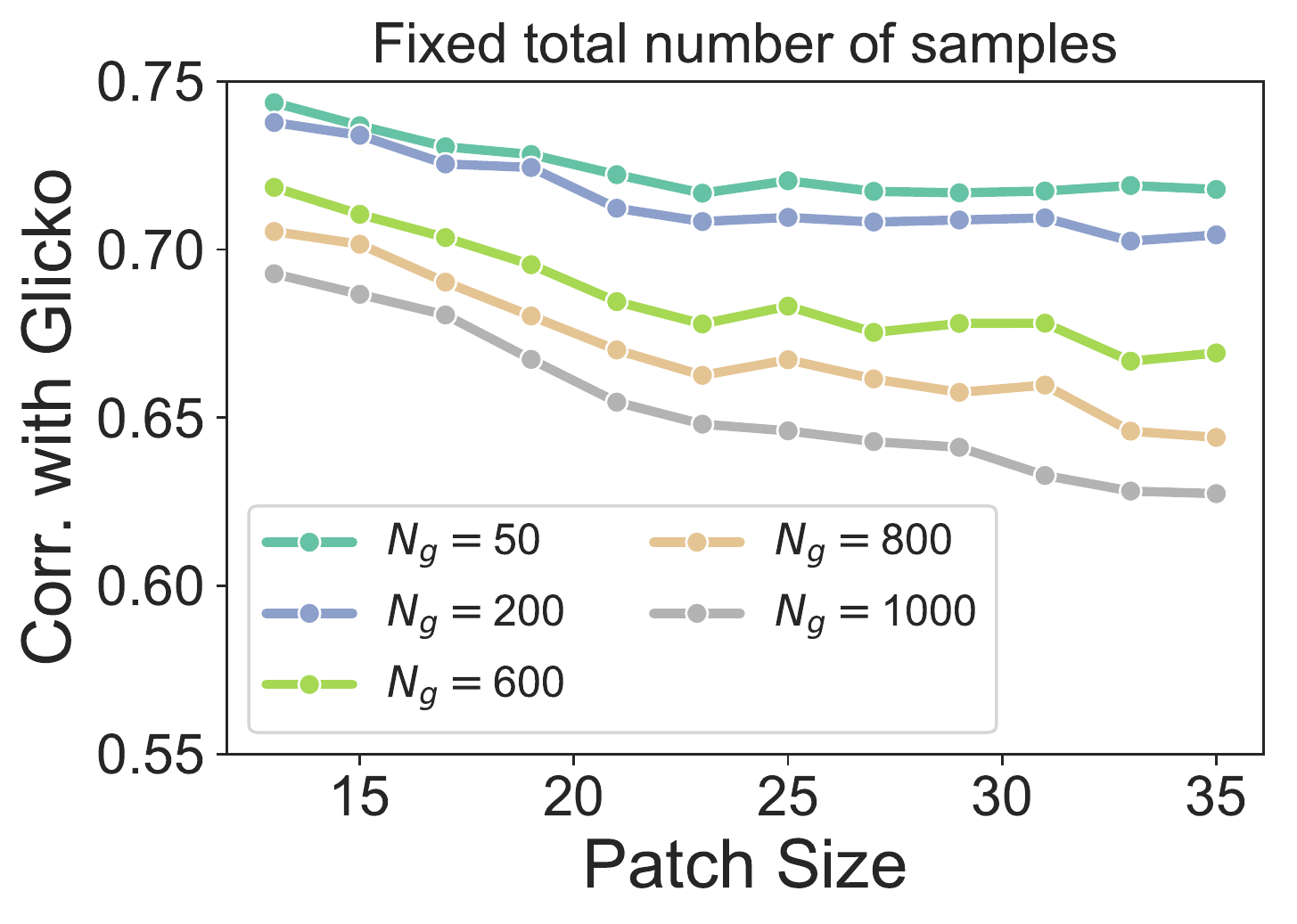} &
\includegraphics[width=0.4\linewidth]{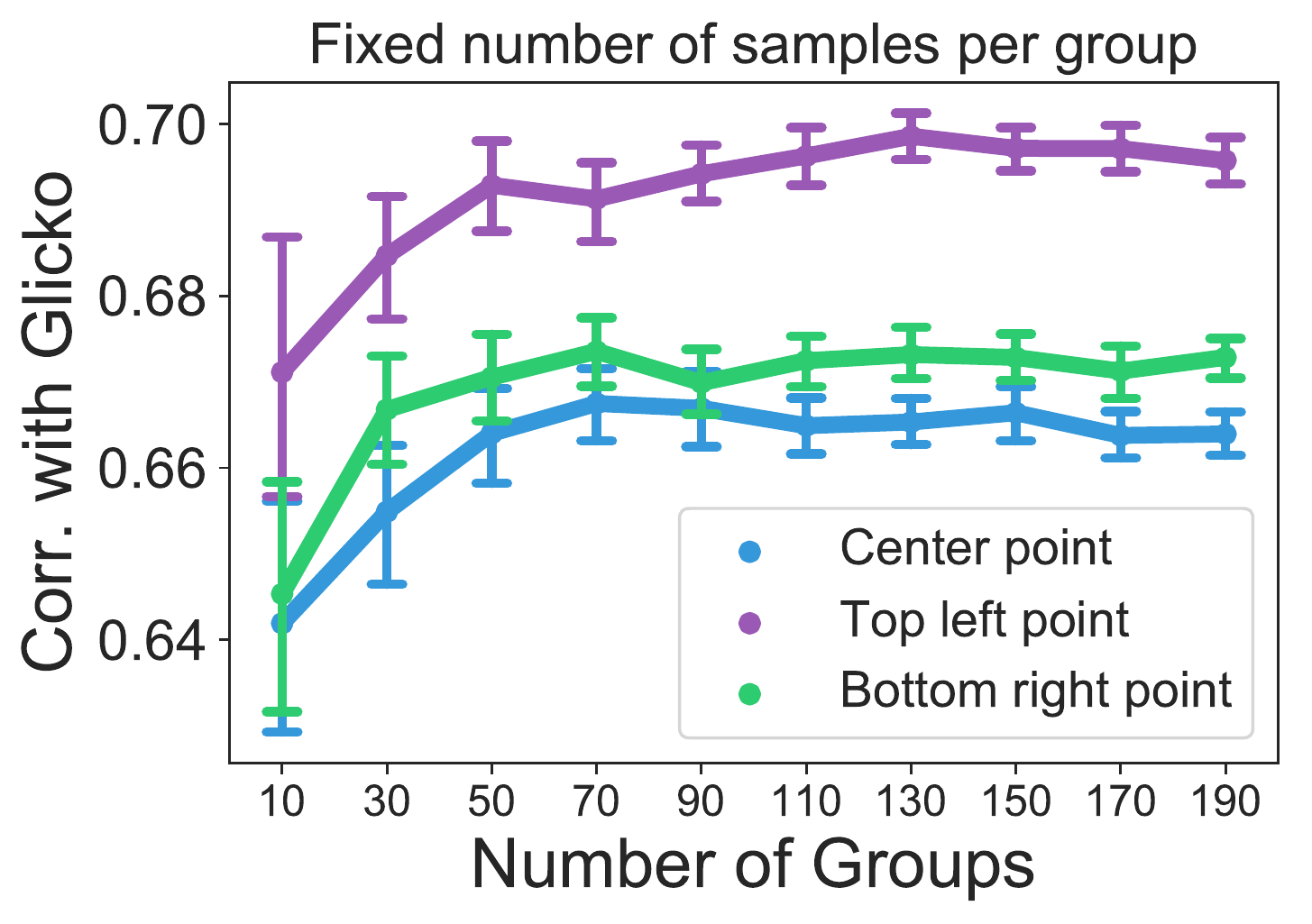} &
\includegraphics[width=0.4\linewidth]{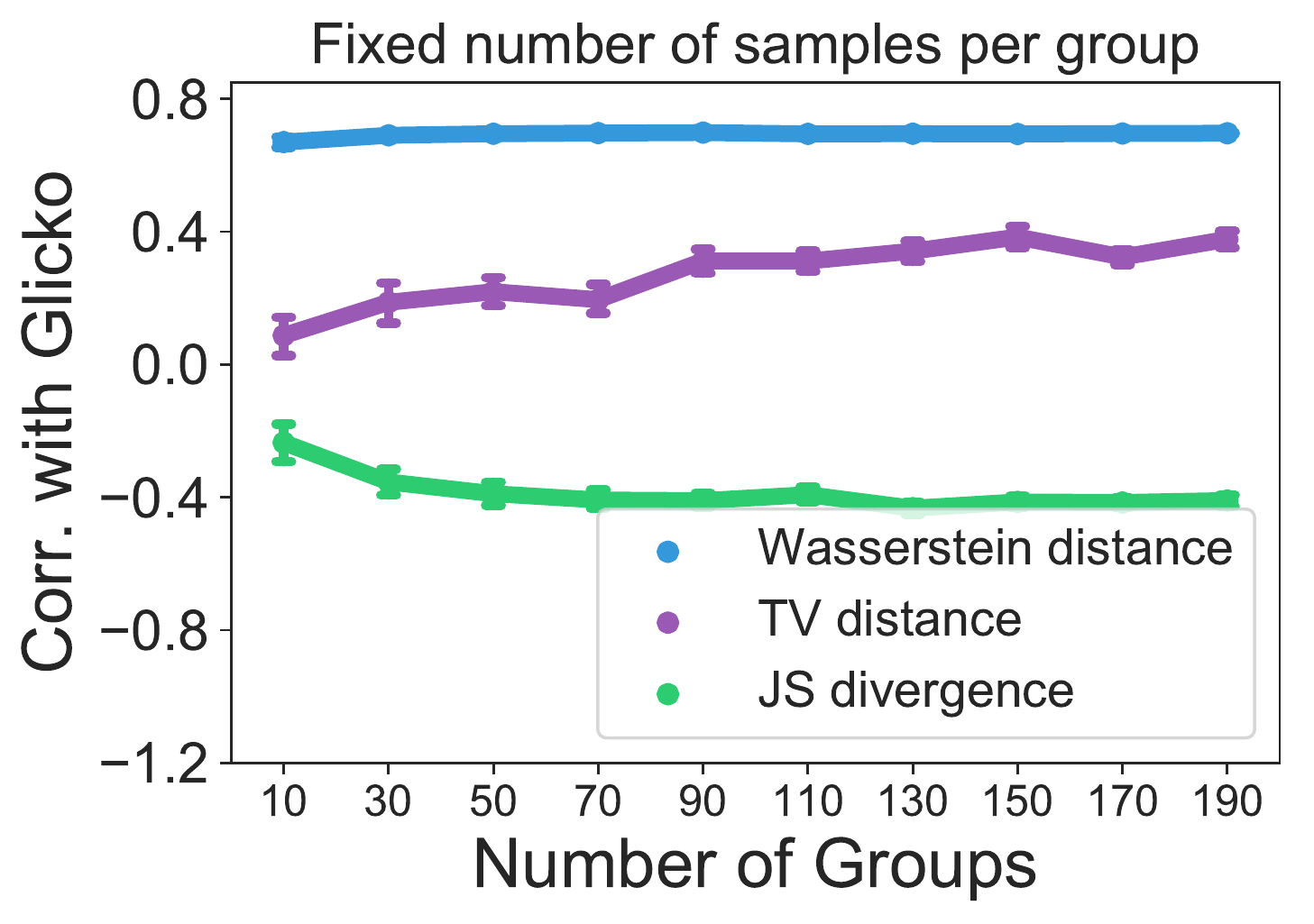} 
 \\
\hspace{4mm} (e) Number of samples per group $N_s$ & \hspace{8mm}(f) LR patch size $r$ &\hspace{7mm} (g) Projection on $Y$ & \hspace{7mm} (h) Distribution distance $d$ \\ \\
 \hline
 \\
\hspace{-3mm}
\includegraphics[width=0.38\linewidth]{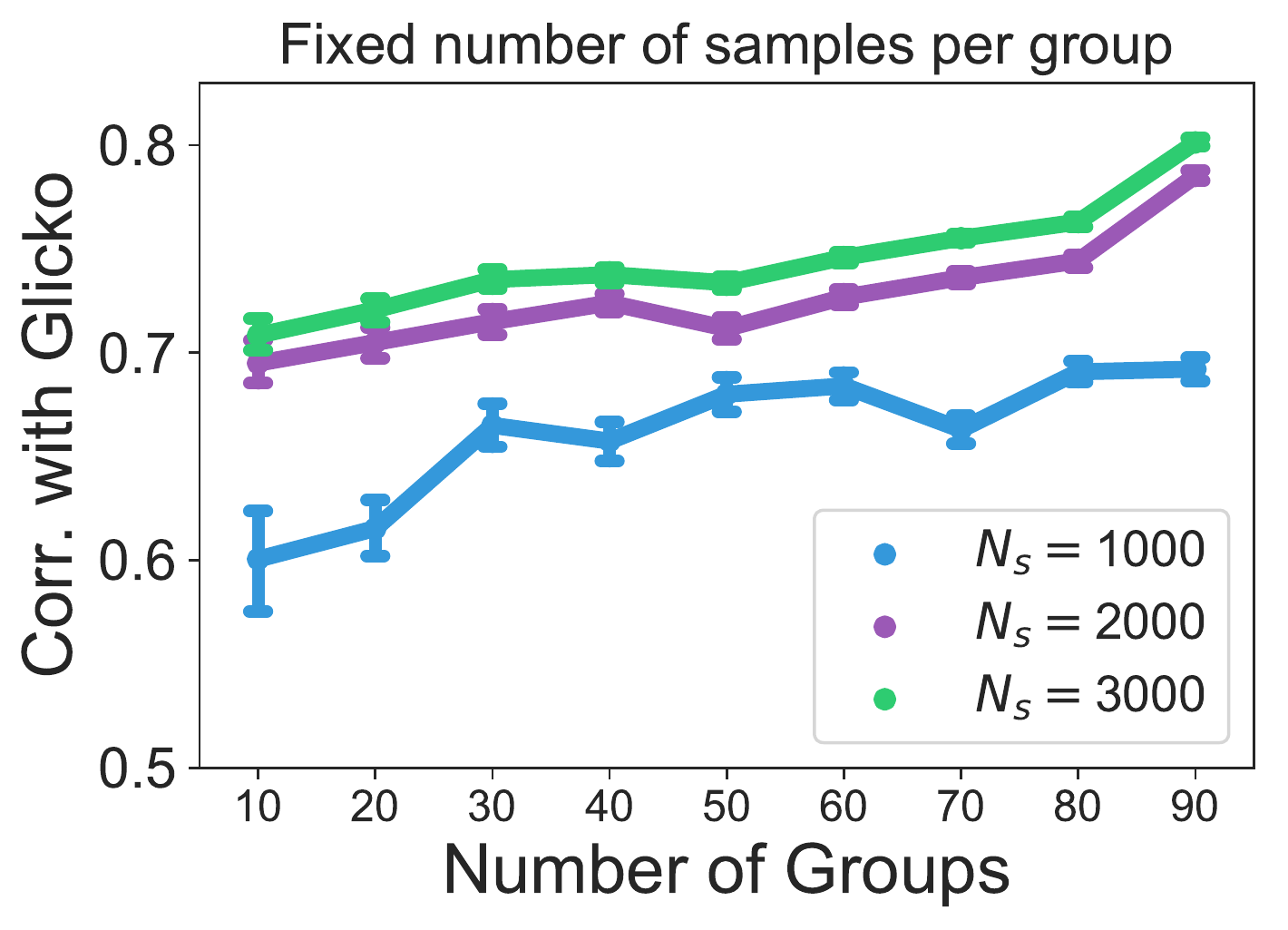} &
\includegraphics[width=0.39\linewidth]{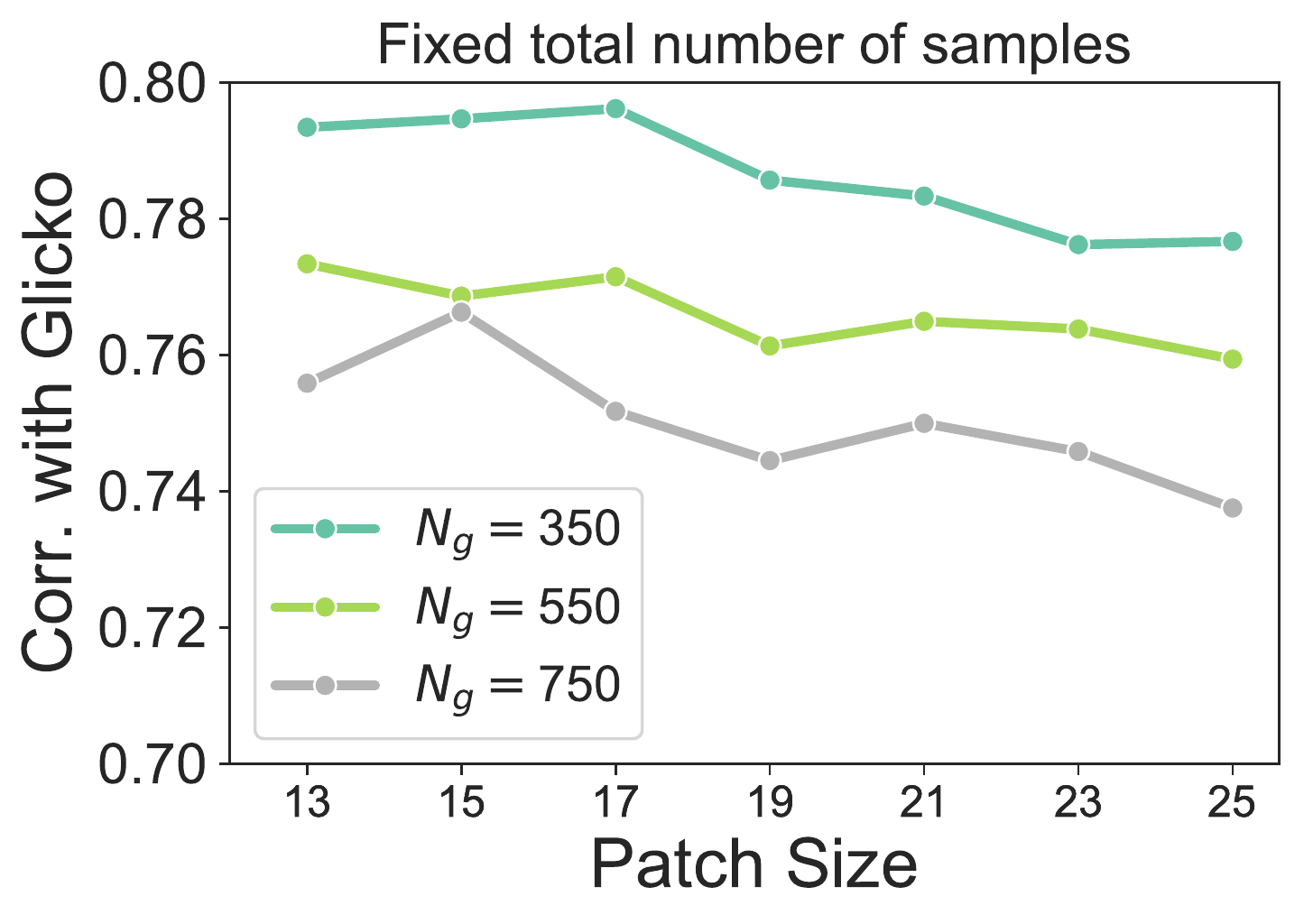} &
\includegraphics[width=0.4\linewidth]{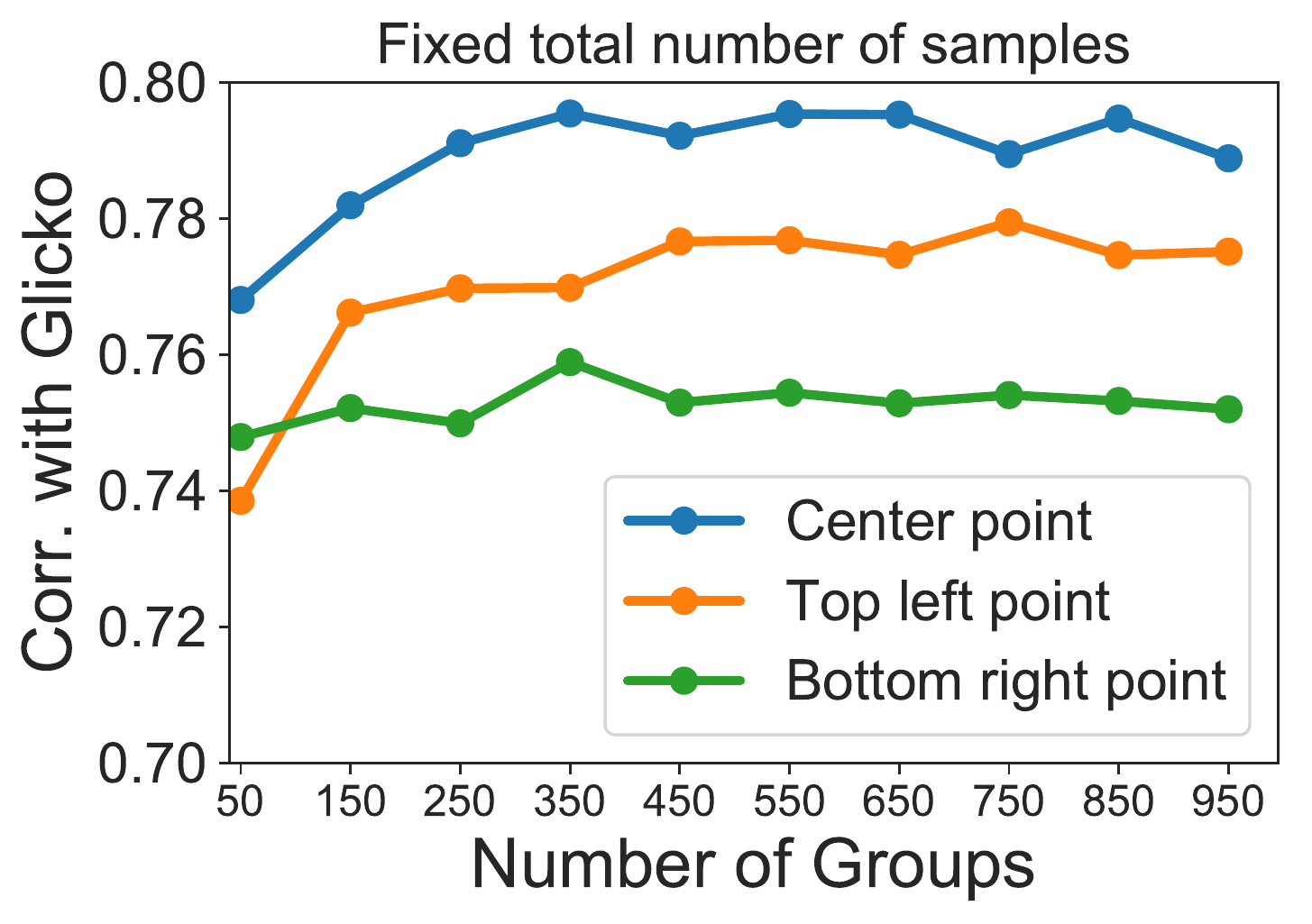} &
\includegraphics[width=0.4\linewidth]{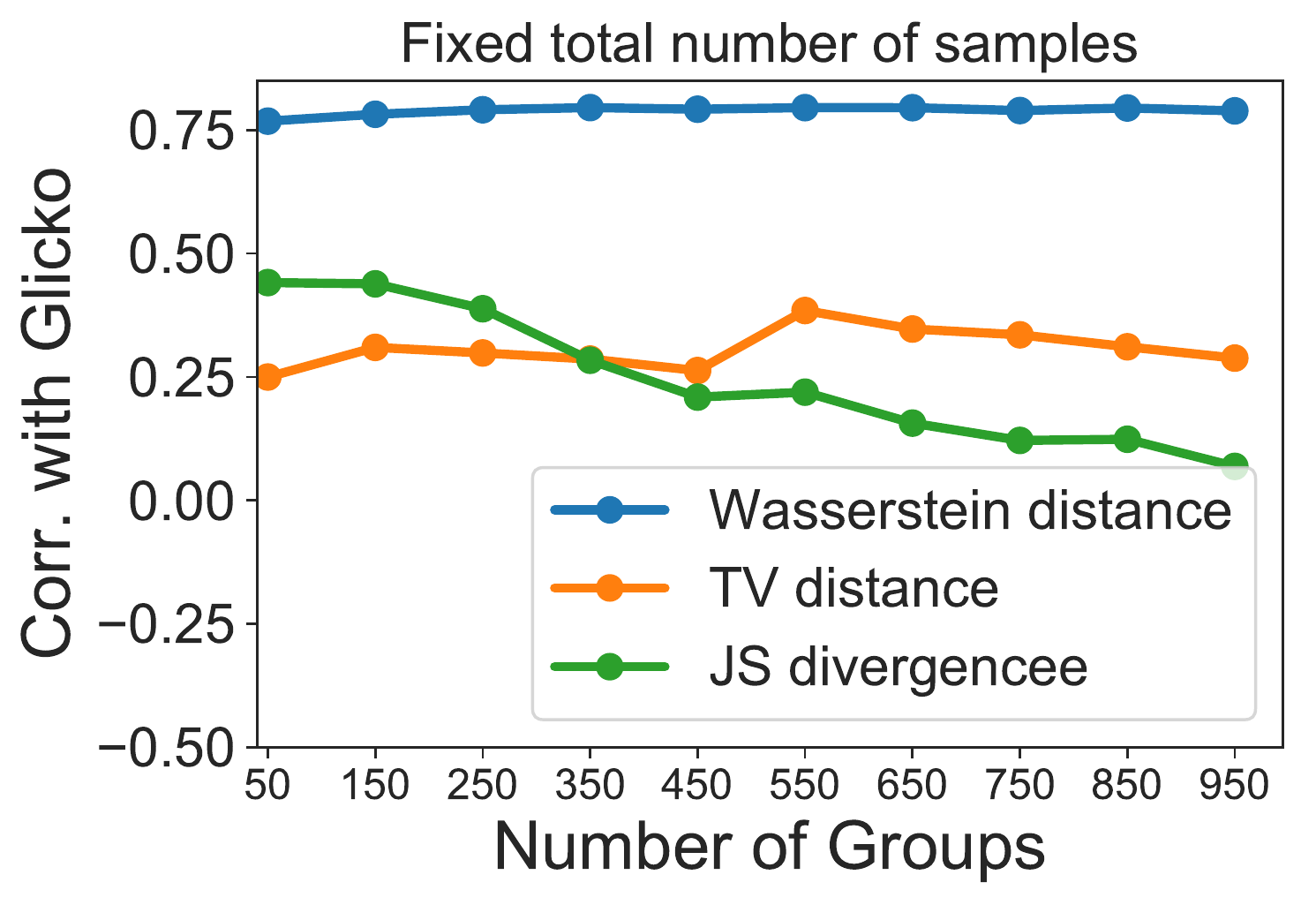} 
 \\
\hspace{4mm} (i) Number of samples per group $N_s$ & \hspace{8mm} (j) LR patch size $r$ & \hspace{8mm} (k) Projection on $Y$ & \hspace{8mm} (l) Distribution distance $d$ \hspace{7mm}\\
\end{tabular}
\end{adjustbox}
\caption{\textbf{Analysis on different design settings in the proposed metric with respect to the correlation with Glicko scores and back-projection error.} 
The first two rows show the results of the proposed metric with grouping on the 1-D projection (SRDM-L), and the third row shows the results of the proposed metric with grouping on original LR patches (SRDM-H).
For SRDM-L, we conduct experiments for each design on two scenarios, I) with a fixed total number of samples, II) with a fixed number of samples per group, while for SRDM-H, we only evaluate on scenario I as it is difficult to obtain sufficient samples for some groups in this case.
(a) and (b) show the impact of different grouping methods on scenario I), on the correlation with Glicko scores and back-projection error, respectively. 
Here we compare grouping methods, SRDM-H (refered as Direct), SRDM-L using first principal component (FPC), and SRDM-L using vector of ones.
(c)(g)(k) and (d)(h)(l) show the impact of different projection choices on $Y$ and distribution distance $d$ on SRDM-L scenario I, SRDM-L scenario II, and SRDM-H scenario I, respectively. 
(e)(i) and (f)(j) show the impact of the number of samples per group $N_s$ and the LR patch size with respect to the number of group $N_g$ on SRDM-L scenario I and SRDM-H scenario I, respectively.
}
\label{fig:desgin}
\end{figure*}

\subsection{Properties of the Proposed Metric}
\label{subsec:design_choice}
In this section, we conduct experiments to verify the design choices of the proposed metric, including the grouping method, the number of groups, the size of the input LR patch, the projection vector for the HR patch, and the distribution distance.
We also show that the proposed metric is robust with respect to the parameters to a certain degree.
If not specified, the default setting described in \secref{subsec:implementation_details} is used.

{\flushleft \bf Grouping method for LR patch.}
As mentioned in \secref{subsec:implementation_details}, we propose two grouping approaches based on the original LR patch space and its projected 1-d space.
We evaluate the correlation between the Glicko scores and the proposed metric with two grouping approaches, as shown in \figref{fig:desgin}(a).
For the projection, we also evaluate two choices, the first principal component (FPC) of LR patches and the vector of ones.
We note that the FPC estimated from a large number of LR patches approximates to a Gaussian matrix, whose elements are larger when closing to the center and roughly centrosymmetric.
The metric with the clustering on the LR patch space obtain high correlation compared with those on the projected 1-d space.
And the clustering with two projection vectors leads to similar results.
\sheng{We further explore the correlation between back-projection error and proposed metric with two grouping method in in Figure~\ref{fig:desgin}(b)}
{\flushleft \bf Number of groups.} 
To understand the impact of the number of groups $N_g$, we show how the correlation between the Glicko scores/back-projection error and the proposed metric varies with respect to $N_g$. 
We first conduct experiments on the situation that the number of samples per group $N_s$ is fixed and $N_g$ is varying.
Among the samples of each group, we randomly sample $N_s$ samples to compute the distribution distance between the estimated results and the ground truth. 
We repeat the process 1000 times and calculate the mean correlation and the variance for each $N_g$.
In \figref{fig:desgin}(e), we show the correlation curves of the Glicko scores with different numbers of samples per group $N_s = 500,1000,2000,3000$. 
The mean correlation increases smoothly with an increasing number of groups for all $N_s$, indicating that finer grouping (samples in each group are more similar to each other) would lead to more accurate measures.
Comparing different curves of $N_s$, larger $N_s$ leads to higher mean correlation and lower variance, as a larger number of samples per group better approximates the distribution ($p_{w^T Y|X \in G}$ and $p_{w^T Y^*|X \in G}$), and therefore gives more accurate distribution distance.

%
We further investigate the practical situation when the number of total LR/HR samples is fixed. 
In this case, when the number of groups $N_g$ increases, the average number of samples per group decreases, and therefore the distribution difference of each group becomes less accurate. 
As shown in \figref{fig:desgin}(a)(``Direct''), the correlation between the Glicko scores and the proposed metric is stable (for all the grouping methods) when $N_g$ is increasing from 50 to 1000.
It is difficult to control the number of samples per group in practice, but we empirically find an average number of samples of 1000 is sufficient for a reasonable result.
{\flushleft \bf Size of LR patch.}
In the proposed metric, the size of the input LR patch would affect the grouping result, and therefore the final result. 
To show its impact, we conduct experiments using different LR patch size $r \in \{13,15,\cdots, 35\}$, and plot the correlation curve between the Glicko scores and the proposed metric. 
As shown in Figure~\ref{fig:desgin}(f)(j), when using smaller LR patch size, the correlation slightly decreases. 
In fact, when we use smaller LR patch size, we group patches with information that is closer to the center pixel (as well as the selected pixel chosen from $w^T Y$), and therefore more center-oriented grouping. 
We choose 13 as the default considering including a sufficient LR region and thus a good context of the region, even when $w^T Y$ selects one pixel. 
%
{\flushleft \bf Projection vector for HR patch.}
As described in \secref{subsec:implementation_details}, we propose to use the projection vector that selects a HR pixel corresponding to the center pixel of the LR patch. 
There are $s \times s$ HR pixels corresponding to that pixel. 
We evaluate the correlation with respect to the choice of the pixel selection, \ie, the center pixel, the top left pixel, and the bottom right pixel.
\figref{fig:desgin}(c) and \figref{fig:desgin}(g) show the results under two scenarios, fixing the total number of samples and fixing the number of samples per group as 500, respectively.
The results using the center pixel is slightly better than the others ($\sim$0.02 in correlation), and the metric is not very sensitive to the choice of the pixel.
Therefore, when there lacks of sufficient samples for computing $p_{w^T Y|X\in G}$ and $p_{w^T Y^*|X\in G}$, it would be a reasonable compromise to approximating them by treating all the pixels in the $s \times s$ region as $w^T Y$/$w^T Y^*$.
%


%
{\flushleft \bf Distribution distance.}
There are a few candidates for the distribution distance, \ie, TV distance, JS divergence, and Wasserstein distance.
For comparison, we present the correlation results using different distribution distances in \figref{fig:desgin}(d)(h) under two scenarios, fixing the total number of samples and fixing the number of samples per group as 500, respectively.
The proposed metric with the Wasserstein distance consistently outperforms the other two, indicating that the Wasserstein distance fits the problem better as also discussed in \secref{subsec:implementation_details}.
Moreover, the metric using the Wasserstein distance is less sensitive to the number of groups when the total number of samples is fixed.

\begin{figure*}[ht]
    \centering
    \begin{adjustbox}{width=0.91\linewidth}
    \hspace{-3mm}
    \centering
    \begin{tabular}[t]{ccc}
        \begin{tabular}{c}
            \includegraphics[width=0.57\linewidth, height=0.44\linewidth]{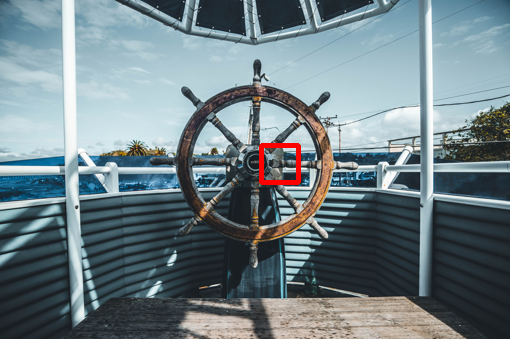}    \\
            (a) LR image
        \end{tabular}
        \hspace{-1.5mm}
        \begin{tabular}{cccc}
            \includegraphics[width=0.2\linewidth]{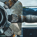} & 
            \includegraphics[width=0.2\linewidth]{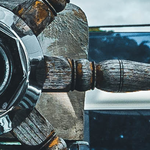} & 
            \includegraphics[width=0.2\linewidth]{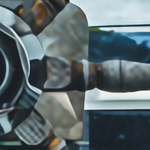} & 
            \includegraphics[width=0.2\linewidth]{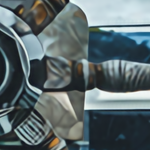} \\
            (b) LR & (c) HR & (d) RCAN~\cite{Zhang_2018_ECCV} & (e) $r$=25, $N_g$=120 \\
            \includegraphics[width=0.2\linewidth]{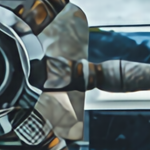} & 
            \includegraphics[width=0.2\linewidth]{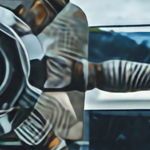} & 
            \includegraphics[width=0.2\linewidth]{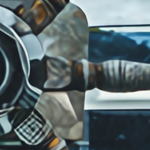} & 
            \includegraphics[width=0.2\linewidth]{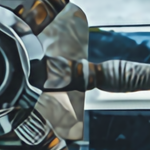} \\
            (f) $r$=19, $N_g$=120 & (g) $r$=13, $N_g$=120 & (h) $r$=25, $N_g$=80 & (i) $r$=25, $N_g$=160
        \end{tabular}
        \vspace{0.2cm}
        \\
        \begin{tabular}{c}
            \includegraphics[width=0.57\linewidth, height=0.44\linewidth]{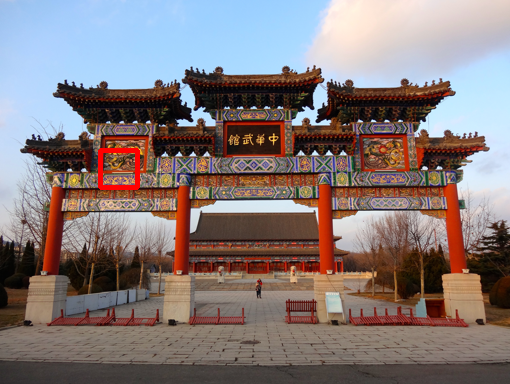}    \\
            (a) LR image
        \end{tabular}
        \hspace{-1.5mm}
        \begin{tabular}{cccc}
            \includegraphics[width=0.2\linewidth]{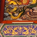} & 
            \includegraphics[width=0.2\linewidth]{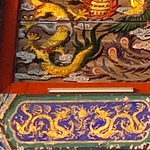} & 
            \includegraphics[width=0.2\linewidth]{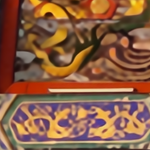} & 
            \includegraphics[width=0.2\linewidth]{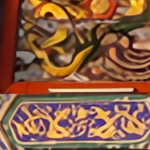} \\
            (b) LR & (c) HR & (d) RCAN~\cite{Zhang_2018_ECCV} & (e) $r$=25, $N_g$=120 \\
            \includegraphics[width=0.2\linewidth]{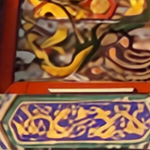} & 
            \includegraphics[width=0.2\linewidth]{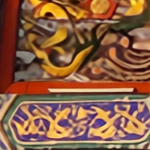} & 
            \includegraphics[width=0.2\linewidth]{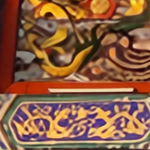} & 
            \includegraphics[width=0.2\linewidth]{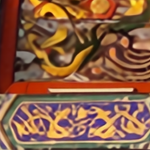} \\
            (f) $r$=19, $N_g$=120 & (g) $r$=13, $N_g$=120 & (h) $r$=25, $N_g$=80 & (i) $r$=25, $N_g$=160
        \end{tabular}
        \vspace{0.2cm}
        \\
        \begin{tabular}{c}
            \includegraphics[width=0.57\linewidth, height=0.44\linewidth]{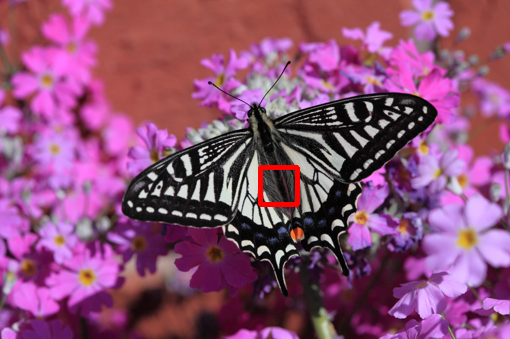}    \\
        (a) LR image
        \end{tabular}
        \hspace{-1.5mm}
        \begin{tabular}{cccc}
            \includegraphics[width=0.2\linewidth]{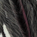} & 
            \includegraphics[width=0.2\linewidth]{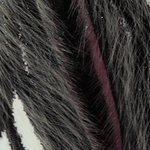} & 
            \includegraphics[width=0.2\linewidth]{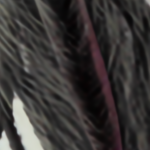} & 
            \includegraphics[width=0.2\linewidth]{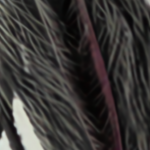} \\
            (b) LR & (c) HR & (d) RCAN~\cite{Zhang_2018_ECCV} & (e) $r$=25, $N_g$=120 \\
            \includegraphics[width=0.2\linewidth]{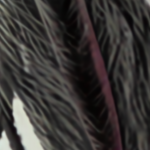} & 
            \includegraphics[width=0.2\linewidth]{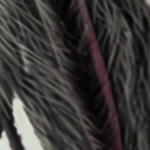} & 
            \includegraphics[width=0.2\linewidth]{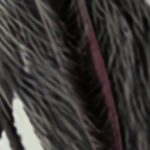} & 
            \includegraphics[width=0.2\linewidth]{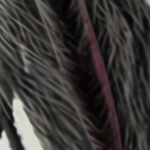} \\
            (f) $r$=19, $N_g$=120 & (g) $r$=13, $N_g$=120 & (h) $r$=25, $N_g$=80 & (i) $r$=25, $N_g$=160
        \end{tabular}
    \end{tabular}
    \end{adjustbox}
\caption{\textbf{Qualitative results on the RCAN~\cite{Zhang_2018_ECCV} models finetuned with the proposed metric.} 
The models are finetuned from the re-trained RCAN model (d).
We investigate the impact of two metric parameters on the network training, the LR patch size $r$ and the number of groups $N_g$, and show the results from models using different settings in (e)-(i).}
\label{fig:training1}
\end{figure*}

\subsection{Network Training with Proposed Metric}
As a metric specifically design for SR problems, an interesting question is: can the proposed metric be used to guide the SR network training?
And the answer is affirmative. 
In the following, we show how to use the proposed metric as the loss to improve a state-of-the-art SR network, RCAN~\cite{Zhang_2018_ECCV}.
To use the proposed metric for SR network training, the major challenge is that the distribution distance used in the metric (\ie, Wasserstein distance) is non-differentiable.
To address the problem, we adopt the sliced Wasserstein distance~\cite{deshpande2018generative,kolouri2019generalized} for $d$ instead of the Wasserstein distance.
The sliced Wasserstein distance $d_{s}$ is an alternative to the Wasserstein distance, and can be easily computed based on sliced one dimensional data, as
\begin{equation}
    d_{s}(p_{w^T Y|X\in G},p_{w^T Y^*|X\in G}) = \sum_{i=1}^{|h_{w^T Y}|} |h_{w^T Y}(\sigma(i)) - h_{w^T Y^*}(\sigma^*(i))|^2_2,
    \label{eq:loss}
\end{equation}
where we denote $h_{w^T Y}$ and $h_{w^T Y^*}$ as the distribution histograms of $w^T Y$ and $w^T Y^*$ respectively.
$i$ represents the index of the histogram, $\sigma$ and $\sigma^*$ are the index permutation functions following the condition:
\begin{equation*}
    h_{w^T Y}(\sigma(i)) \leqslant h_{w^TY}(\sigma(i+1)), \quad h_{w^TY^*}(\sigma^*(i)) \leqslant h_{w^TY^*}(\sigma^*(i+1)).
\end{equation*}

For SR network training, we adopt the metric with the grouping approach on the projected space, SRDM-L, as mentioned in \secref{subsec:implementation_details}.
The loss function is defined based on \eqref{eq: second metric} with the sliced Wasserstein distance, and is calculated on the LR/HR patches from a batch of LR/HR images.
For each training batch, there lacks of enough samples to approximate the distribution $p_{w^T Y|X\in G}$ and $p_{w^T Y^*|X\in G}$ for each group $G$. 
Therefore, we treat all the pixels in the $s \times s$ region, corresponding to the center pixel of the LR patch, the same for $w^T Y$/$w^T Y^*$, as discussed in \secref{subsec:design_choice}.
Moreover, for a good evaluation of the distribution based metric, we propose to use more but smaller LR/HR samples in a training batch, to include sufficient variety of LR patches. 
In this work, we use 400 LR images of size $36 \times 36$ in a training batch.

The training process of each iteration can be summarized as below,
\begin{itemize}
    \item Collect LR/HR image patches through sliding windows, and group them on the projected space using the FPC as described in \secref{subsec:implementation_details}. 
    \item Collect the statistics of $p_{w^T Y|X\in G}$ and $p_{w^T Y^*|X\in G}$ for each group $G$.
    \item Compute loss using \eqref{eq: second metric} with the sliced Wasserstein distance for backpropagation.
\end{itemize}
The main issue to use the proposed metric for SR network training is the expensive computational load, as calculating the loss requires collecting the statistics for the distribution $p_{w^T Y|X\in G}$ and $p_{w^T Y^*|X\in G}$. 
Each training iteration takes around 120 seconds for a batch on a machine with INTEL Xeon 2.40GHz and four NVIDIA P100 (24GB).
In this case, it would be very time-consuming to train a network from the scratch.
%
%
Therefore, we advocate using the metric to finetune the pre-trained network with the $\ell_1$/$\ell_2$ loss.
In this work, we consider finetuning the RCAN~\cite{Zhang_2018_ECCV} model trained using the $\ell_1$ loss. 
And we use the code and the trained model provided by the authors.

We conduct experiments on the DIV2K~\cite{timofte2018ntire} dataset, by finetuning the model on the training set and evaluating on the test set. 
The model is finetuned until convergence, and it takes around 800 iterations. 
In \figref{fig:training1}, we provide some qualitative comparisons on the pretrained RCAN and the finetuned RCAN using the proposed metric.

Moreover, we investigate the impact of two metric parameters on the network training, the number of groups $N_g$ and the LR patch size $r$.
Both the LR patch size and the number of groups would influence the grouping results of LR patches.
To evaluate the impact of the LR patch size, we fix the number of groups as 120 (to maintain a sufficient average number of samples per group in a batch), and test patch sizes $r \in \{13,19,25\}$.
The network with a smaller $r$ tends to generate sharper images with more details, as shown in \figref{fig:training1}(e)-(g).
The reason could be that smaller $r$ in the metric indicates using smaller regions to determine the $s \times s$ HR region corresponding to the center pixel of LR patch, thus the network would generate finer details based on more local textures.
However, some of the textures are synthesized and should not exist based on the global context.
To evaluate the impact of the number of groups, we fix the LR patch size as 25 (to maintain a sufficient average number of samples per group in a batch), and the number of groups $N_g \in \{80,120,160\}$.
The network with a larger $N_g$ tends to generate more subtle results, as a larger $N_g$ leads to finer grouping and therefore better distribution estimation. 
However, if the $N_g$ is too large, the average number of samples per group would be small, and could result in groups with insufficient samples for accurate distribution estimation.
Considering the analysis above, we consider $r=25,N_g=160$ as a good setting for this experiments.

We further show the quantitative comparison between the pretrained RCAN and the refined RCAN using the proposed metric, with respect to PSNR/SSIM/Ma/LPIPS, in \tabref{tab:training_table}.
The refined RCAN achieves significant improvement in terms of perceptual quality (\ie, Ma and LPIPS) while sacrifices a bit fidelity (\ie, PSNR/SSIM), which matches the analysis in \secref{subsec:RelationtoPQ} and \secref{subsec:RelationtoSR}.
Due to limited space, more experimental results and analysis will be provided in the supplementary material.

\section{Conclusion}
In this work, We propose a new distribution-based metric for super resolution based on its one-to-many nature.
We show that the proposed metric is highly correlated to the perceptual quality, and more correlated to the perceptual quality based metrics, \eg, LPIPS and Ma. 
We conduct extensive experiments to show the property of the proposed metric, and show that it is robust to the parameter setting.
Moreover, we demonstrate that the proposed metric can be used for training SR networks for better visual results.


\clearpage
%
%

\bibliographystyle{splncs04}
\bibliography{egbib}

\appendix

\section{Overview}
In this supplementary material, we present additional training results using the proposed metric. 
First, we show more examples of the finetuned RCAN model using the proposed metric on the B100 dataset~\cite{MartinFTM01}. 
Second, we show that improving SR networks using the proposed metric works for other models. 
Specifically, we test the EDSR~\cite{lim2017enhanced} model and provide results on the DIV2K dataset~\cite{timofte2018ntire}. 

\section{Addition Results on Finetuning RCAN}
\begin{figure}[!h]
    \centering
    \begin{adjustbox}{width=\linewidth}
    \hspace{-3mm}
    \centering
    \begin{tabular}[t]{cc}
        \begin{tabular}{c}
            \begin{tabular}{c}
                \includegraphics[width=0.72\linewidth]{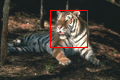}    \\
                (a) LR image \\
            \end{tabular}\\
            \begin{tabular}{cc}                                                              \includegraphics[width=0.35\linewidth]{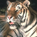} &\hspace{1mm}
                \includegraphics[width=0.35\linewidth]{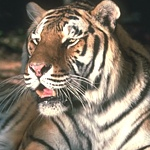} 
                \\
                (b) LR  &(c) HR \\
            \end{tabular} \\
            \begin{tabular}{cc}
                \includegraphics[width=0.35\linewidth]{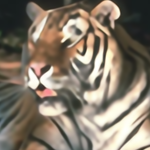}  &\hspace{1mm}
                \includegraphics[width=0.35\linewidth]{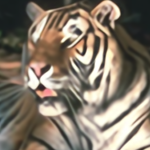} \\
                 (d) RCAN~\cite{Zhang_2018_ECCV} &  (e) $r$=25, $N_g$=120\\
                \includegraphics[width=0.35\linewidth]{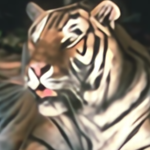} & \hspace{1mm}
                \includegraphics[width=0.35\linewidth]{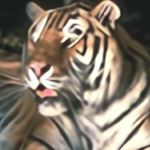} \\
                (f)  $r$=19, $N_g$=120 & (g) $r$=13, $N_g$=120 \\
                \includegraphics[width=0.35\linewidth]{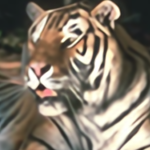} & \hspace{1mm}
                \includegraphics[width=0.35\linewidth]{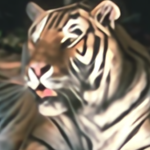} \\
                 (h) $r$=25, $N_g$=80 & (f) $r$=25, $N_g$=160 \\
                
            \end{tabular}
        \end{tabular}
        \hspace{0.3cm}
        &
        \begin{tabular}{c}
            \begin{tabular}{c}
                \includegraphics[width=0.72\linewidth]{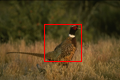}    \\
                (a) LR image \\
            \end{tabular}\\
            \begin{tabular}{cc}
                 \includegraphics[width=0.35\linewidth]{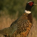}& \hspace{1mm}
                \includegraphics[width=0.35\linewidth]{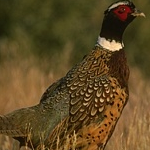} 
                \\
                (b) LR  &(c) HR \\
            \end{tabular} \\
            \begin{tabular}{cc}
                \includegraphics[width=0.35\linewidth]{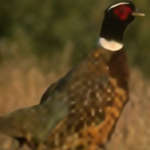}  &\hspace{1mm}
                \includegraphics[width=0.35\linewidth]{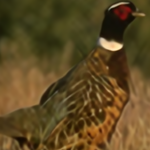} \\
                 (d) RCAN~\cite{Zhang_2018_ECCV} &  (e) $r$=25, $N_g$=120\\
                \includegraphics[width=0.35\linewidth]{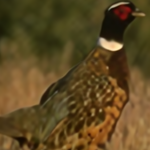} & \hspace{1mm}
                \includegraphics[width=0.35\linewidth]{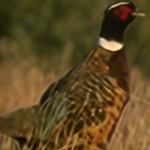} \\
                (f)  $r$=19, $N_g$=120 & (g) $r$=13, $N_g$=120 \\
                \includegraphics[width=0.35\linewidth]{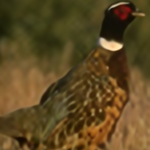} & \hspace{1mm}
                \includegraphics[width=0.35\linewidth]{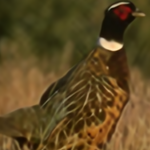} \\
                 (h) $r$=25, $N_g$=80 & (f) $r$=25, $N_g$=160 \\
                
            \end{tabular}
        \end{tabular}
        
    \end{tabular}
    \end{adjustbox}
\caption{\textbf{Qualitative results on finetuning the RCAN~\cite{Zhang_2018_ECCV} model with the proposed metric on B100~\cite{MartinFTM01} dataset.} 
The models in (e)-(f) are finetuned from the pretrained RCAN model (d) on B100.
We investigate the impact of two metric parameters on the network training, the LR patch size $r$ and the number of groups $N_g$, and show the results from models using different settings in (e)-(f).}
\label{fig:training_b100}
\end{figure}

Other than the results of the finetuned RCAN on DIV2K dataset in the manuscript, we test finetuning the RCAN model on the B100 dataset~\cite{MartinFTM01}.
In Figure~\ref{fig:training_b100}, we provide the results with respect to two metric parameters, the LR patch size $r$ and the number of group $N_g$.
%
The finetuned RCAN model obtains sharper results with more details compared with the pretrained model.
%

\section{Addition Results on Finetuning EDSR}

\begin{figure}[!h]
    \centering
    \begin{adjustbox}{width=\linewidth}
    \hspace{-3mm}
    \centering
    \begin{tabular}[t]{cc}
        \begin{tabular}{c}
            \begin{tabular}{c}
                \includegraphics[width=0.72\linewidth]{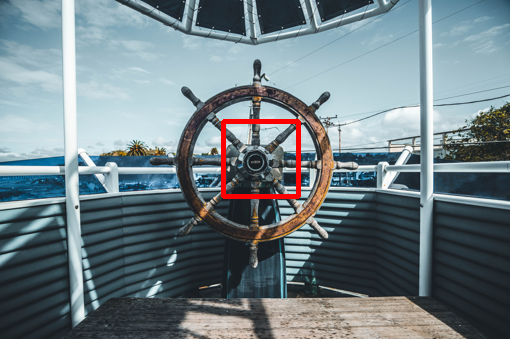}    \\
                (a) LR image \\
            \end{tabular}\\
            \begin{tabular}{cc}                                                                   \includegraphics[width=0.35\linewidth]{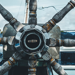} &\hspace{1mm}
                \includegraphics[width=0.35\linewidth]{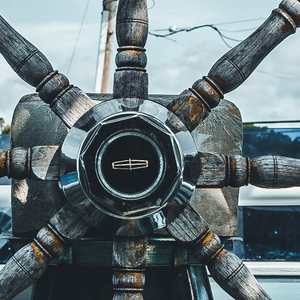} 
                \\
                (b) LR  &(c) HR \\
            \end{tabular} \\
            \begin{tabular}{cc}
                \includegraphics[width=0.35\linewidth]{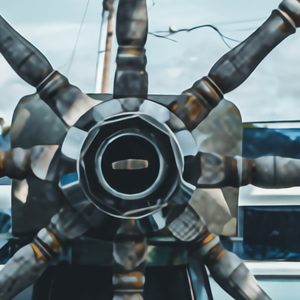}  &\hspace{1mm}
                \includegraphics[width=0.35\linewidth]{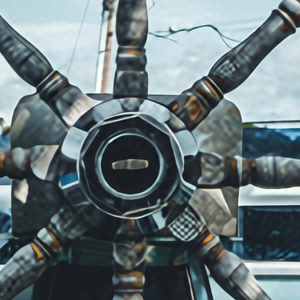} \\
                 (d) EDSR~\cite{lim2017enhanced} &  (e) Ours:  $r$=13, $N_g$=10\\
                
            \end{tabular}
        \end{tabular}
        \hspace{0.3cm}
        &
        \begin{tabular}{c}
            \begin{tabular}{c}
                \includegraphics[width=0.72\linewidth, height = 0.48\linewidth]{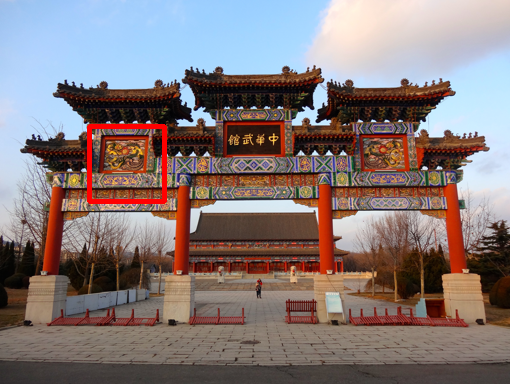}    \\
                (a) LR image \\
            \end{tabular}\\
            \begin{tabular}{cc}                                                                   \includegraphics[width=0.35\linewidth]{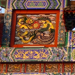} &\hspace{1mm}
                \includegraphics[width=0.35\linewidth]{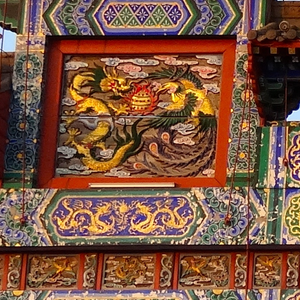} 
                \\
                (b) LR  &(c) HR \\
            \end{tabular} \\
            \begin{tabular}{cc}
                \includegraphics[width=0.35\linewidth]{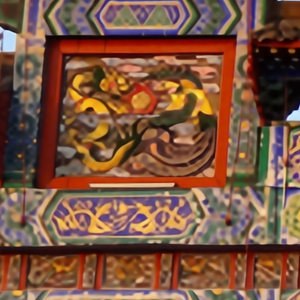}  &\hspace{1mm}
                \includegraphics[width=0.35\linewidth]{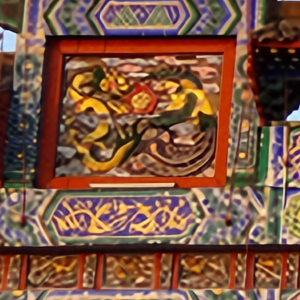} \\
                 (d) EDSR~\cite{lim2017enhanced} &  (e) Ours:  $r$=13, $N_g$=10\\
                
            \end{tabular}
        \end{tabular}
    \end{tabular}
    \end{adjustbox}
\label{fig:training2}
\end{figure}    
        
\begin{figure}[!h]
    \centering
    \begin{adjustbox}{width=\linewidth}
    \hspace{-3mm}
    \centering
    \begin{tabular}[t]{cc}
        \begin{tabular}{c}
            \begin{tabular}{c}
                \includegraphics[width=0.72\linewidth, height = 0.48\linewidth]{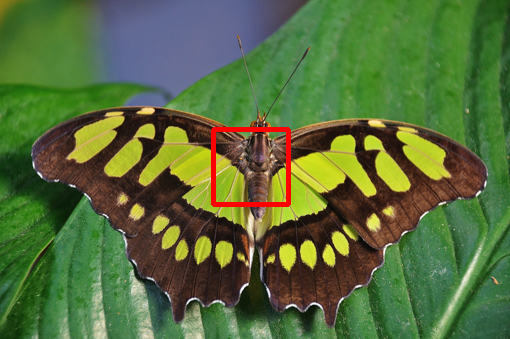}    \\
                (a) LR image \\
            \end{tabular}\\
            \begin{tabular}{cc}                                                                   \includegraphics[width=0.35\linewidth]{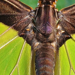} &\hspace{1mm}
                \includegraphics[width=0.35\linewidth]{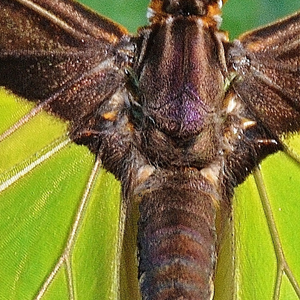} 
                \\
                (b) LR  &(c) HR \\
            \end{tabular} \\
            \begin{tabular}{cc}
                \includegraphics[width=0.35\linewidth]{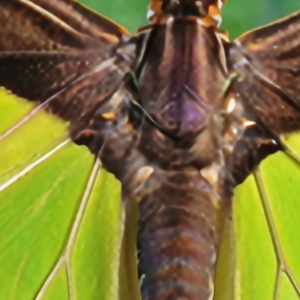}  &\hspace{1mm}
                \includegraphics[width=0.35\linewidth]{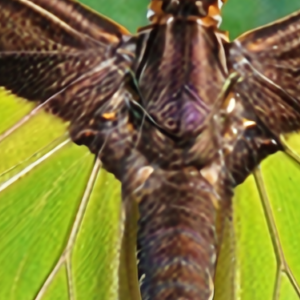} \\
                 (d) EDSR~\cite{lim2017enhanced} &  (e) Ours:  $r$=13, $N_g$=10\\
                
            \end{tabular}
        \end{tabular}
        \hspace{0.3cm}
        &
        \begin{tabular}{c}
            \begin{tabular}{c}
                \includegraphics[width=0.72\linewidth, height = 0.48\linewidth]{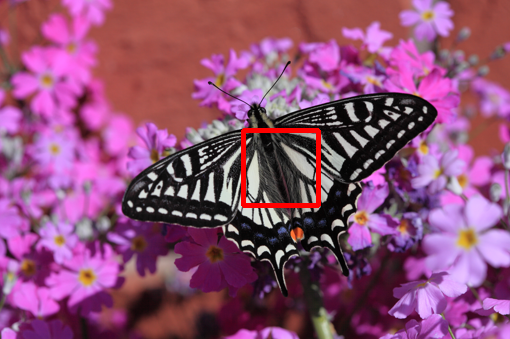}    \\
                (a) LR image \\
            \end{tabular}\\
            \begin{tabular}{cc}                                                                   \includegraphics[width=0.35\linewidth]{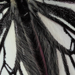} &\hspace{1mm}
                \includegraphics[width=0.35\linewidth]{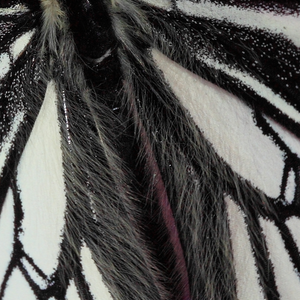} 
                \\
                (b) LR  &(c) HR \\
            \end{tabular} \\
            \begin{tabular}{cc}
                \includegraphics[width=0.35\linewidth]{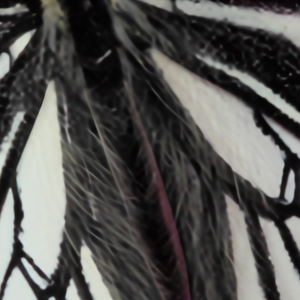}  &\hspace{1mm}
                \includegraphics[width=0.35\linewidth]{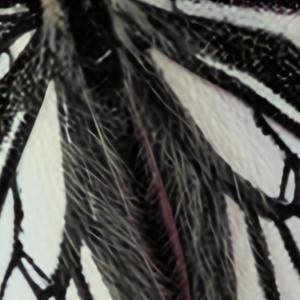} \\
                 (d) EDSR~\cite{lim2017enhanced} &  (e) Ours:  $r$=13, $N_g$=10\\
                
            \end{tabular}
        \end{tabular} 
        
    \end{tabular}
    \end{adjustbox}
\caption{\textbf{Qualitative results on the EDSR~\cite{lim2017enhanced} model finetuned with the proposed metric on the DIV2K dataset~\cite{timofte2018ntire}.} 
The model in (e) is finetuned using the proposed metric from the pretrained EDSR model (d) on the DIV2K dataset.
}
\label{fig:training_edsr}
\end{figure}
To show that the proposed metric works for other models, we test finetuning the EDSR~\cite{lim2017enhanced} model (pretrained using $\ell_1$ loss) using the proposed metric on the DIV2K dataset~\cite{timofte2018ntire}.
To democratize the training using the proposed metric, we conduct the experiment using an economic setup with an INTEL i7 3.70GHz CPU and a NVIDIA 1080Ti (11G RAM) GPU. 
Due to the limited RAM space, we use 100 LR images of size $24 \times 24$ in a batch, and set the metric parameters as $r = 13, N_g = 10$ to maintain similar average samples per group as that in the RCAN experiment of the manuscript. 
%
In Figure~\ref{fig:training_edsr}, we present the results of the pretrained EDSR model using $\ell_1$ loss and the finetuned EDSR model using the proposed metric.
Even when sacrificing the accuracy of the proposed metric, the finetuned EDSR model still achieves better qualitative performance in terms of more textures and sharper results, compared with the pretrained model.


\end{document}